\documentclass[10pt, journal, compsoc]{IEEEtran}

\usepackage{cite}
\usepackage{amsmath,amssymb,amsfonts}
\usepackage{algorithmic}
\usepackage{graphicx}
\usepackage{textcomp}
\usepackage{url}
\usepackage{xcolor}
\usepackage{times}
\usepackage{epsfig}
\usepackage{amssymb}
\usepackage{listings}
\usepackage{booktabs} 
\usepackage{multirow}
\usepackage{gensymb}
\usepackage{mathtools}
\usepackage{adjustbox}
\usepackage{framed}
\usepackage{latexsym}
\usepackage{amsfonts}
\usepackage{dsfont}
\usepackage{float}
\usepackage{tabularx}

\usepackage{arydshln}
\usepackage{xcolor}
\definecolor{DarkGreen}{rgb}{0.2,0.5,0.2} 

\makeatletter
\AtBeginDocument{\def\@citecolor{DarkGreen}}
\makeatother

\usepackage{arydshln}
\newcommand{\etal}{et al.}

\usepackage{amsmath}
\usepackage{graphicx}
\usepackage{tabularx}
\usepackage{color, colortbl}
\usepackage{tikz}
\usepackage{ragged2e}

\usepackage[pagebackref=true,breaklinks=true,colorlinks,bookmarks=false, urlcolor=DarkGreen, citecolor=DarkGreen]{hyperref}

\begin{document}

\title{AffectNet+: A Database for Enhancing Facial Expression Recognition with Soft-Labels}

\author{Ali~Pourramezan~Fard*, Mohammad~Mehdi~Hosseini*\thanks{* These authors contributed equally to this work.}, \IEEEmembership{Student Member, IEEE}, Timothy~D.~Sweeny, and Mohammad~H.~Mahoor, \IEEEmembership{Senior Member, IEEE}

\IEEEcompsocitemizethanks{
    \IEEEcompsocthanksitem Ali Pourramezan Fard, Mohammad Mehdi Hosseini, and Mohammad H. Mahoor are with the Ritchie School of Engineering and Computer Science, University of Denver, Denver,
    CO, 80208 .\protect\\
    E-mails: \{Ali.Pourramezanfard, MohammadMehdi.Hosseini, Mohammad.Mahoor\}@du.edu

    \IEEEcompsocthanksitem Timothy D. Sweeny is with the College of Arts, Humanities and Social Sciences, University of Denver, Denver,
    CO, 80208 .\protect\\
    E-mail: timothy.sweeny@du.edu
    }
}

\markboth{}{Pourramezan Fard \MakeLowercase{\textit{et al.}}: AffectNet+}

\IEEEtitleabstractindextext{%
\begin{abstract}
\label{SEC:ABSTRACT}
\justifying
Automated Facial Expression Recognition (FER) is challenging due to intra-class variations and inter-class similarities. FER can be especially difficult when facial expressions reflect a mixture of various emotions (aka compound expressions). Existing FER datasets, such as AffectNet, provide discrete emotion labels (\textit{hard-labels}), where a single category of emotion is assigned to an expression. To alleviate inter- and intra-class challenges, as well as provide a better facial expression descriptor, we propose a new approach to create FER datasets through a labeling method in which an image is labeled with more than one emotion (called \textit{soft-labels}), each with different confidences. Specifically, we introduce the notion of \textit{soft-labels} for facial expression datasets, a new approach to affective computing for more realistic recognition of facial expressions. To achieve this goal, we propose a novel methodology to accurately calculate \textit{soft-labels}: a vector representing the extent to which multiple categories of emotion are simultaneously present within a single facial expression. Finding smoother decision boundaries, enabling multi-labeling, and mitigating bias and imbalanced data are some of the advantages of our proposed method. Building upon AffectNet, we introduce AffectNet+, the next-generation facial expression dataset. This dataset contains \textit{soft-labels}, three categories of data complexity subsets, and additional metadata such as age, gender, ethnicity, head pose, facial landmarks, valence, and arousal. AffectNet+ will be made publicly accessible to researchers.
\end{abstract}

}

\maketitle

\renewcommand{\thesection}{\arabic{section}}
\renewcommand{\thetable}{\arabic{table}}
\renewcommand{\thefigure}{\arabic{figure}}

\section{Introduction}
\label{SEC:INTRODUCTION}
Facial expressions are essential non-verbal communication channels utilized by both humans and animals~\cite{darwin1998expression}. Facial expressions result from facial muscle movements and provide a window into the emotions, feelings, and psychological states humans experience~\cite{kret2015emotional}. The discrete/categorical theory of emotions defines six basic (potentially universally shared) emotions expressed by facial expressions Happy, Sad, Surprise, Fear, Disgust, and Anger~\cite{ekman1971constants, ekman1999basic}. Contempt, which is the feeling of dislike for and superiority (usually morally) over another person, was later added to this list of basic emotions \cite{ekman1987universals}. Recognition and analysis of emotional facial expressions have many applications including emotion regulation, cultural influences, health care, and human-computer interaction (HCI). While manual measurement of facial expressions is a labor-intensive task, the development of automated Facial Expression Recognition (FER) using machine learning (ML) algorithms has garnered significant attention in the realms of computer vision over the past few decades. Considerable FER advancements have been made in recent years by employing robust deep learning methods, such as Convolutional Neural Networks (CNNs)~\cite{farzaneh2021facial, fard2022ad, hasani2020breg}, and Vision-Transformers~\cite{yang2022face, xue2022vision}. Specifically, in comparison with the traditional ML methods, deep learning-based models have better success in dealing with images collected in uncontrolled environments (\textit{aka} wild settings) where we can witness a vast variation in scene lighting, camera view, image resolution, and subject's head pose, gender, and ethnicity. 

Creating a robust and accurate FER model using machine learning necessitates a substantial dataset of annotated facial images. Annotating facial expressions in images poses challenges due to intrinsic intra-class variations and inter-class similarities~\cite{fard2022ad} among facial expressions. Intra-class variations reflect the diverse range of expressions observed within a single emotion category. For example, sadness can manifest to various degrees with distinct facial muscle movements~\cite{ekman1978facial}. Similarly, happiness can be perceived across a range of different smiles (\textit{e.g.,} Duchenne smile vs non-Duchenne smile~\cite{de1990mechanism}). Inter-class similarities refer to the overlap in activation of facial musculature across different emotion categories, especially evident in subtle expressions. For instance, the high correlation between muscle movements associated with Happy and Contempt expressions causes confusion when distinguishing subtle variations between these emotions.

\begin{figure}[t]
  \centering
  \includegraphics[width=0.8\columnwidth]{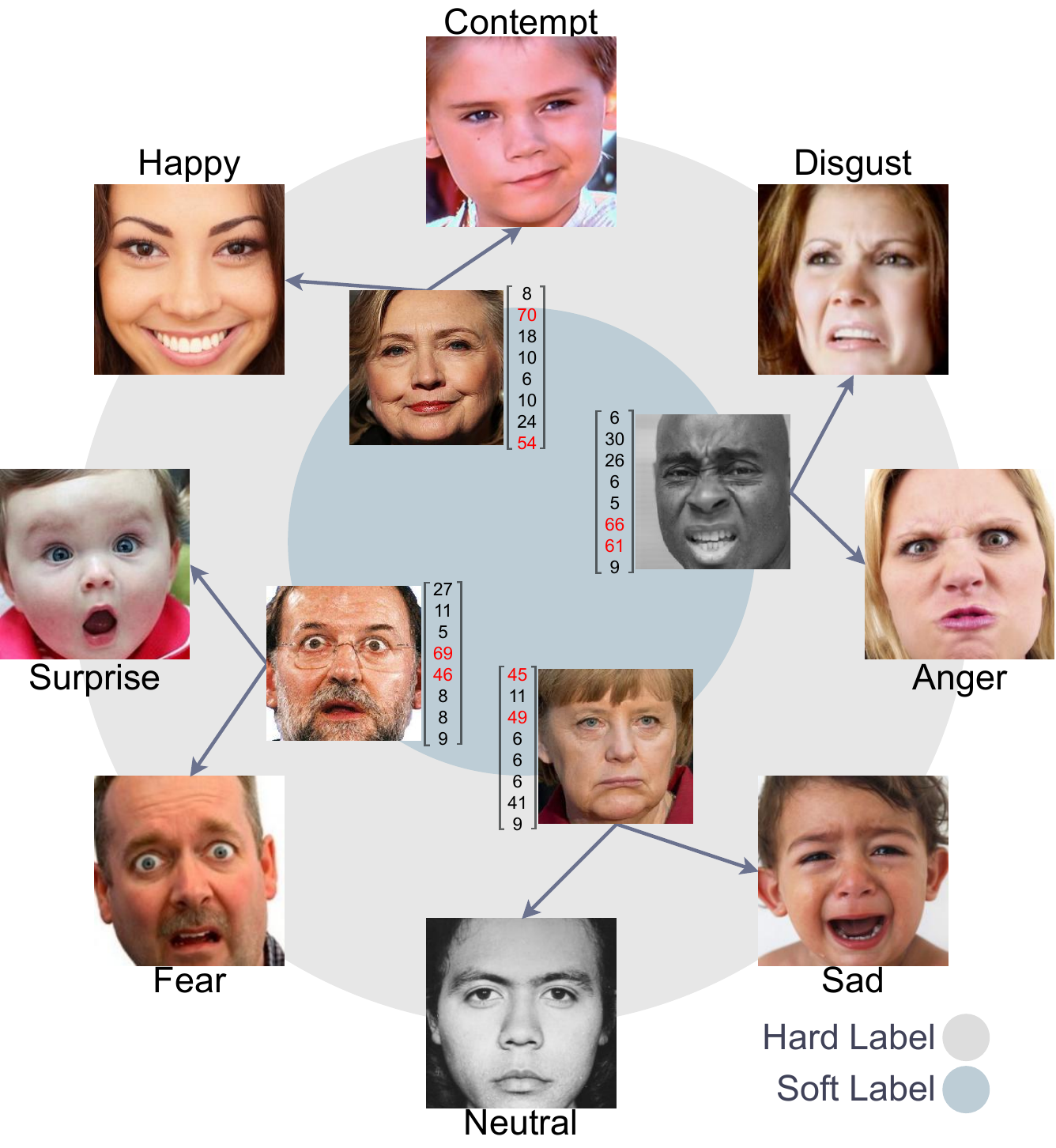}
  \caption{Unlike the traditional approaches, where a single emotion label is assigned to each image, we introduce \textit{soft-labels} to provide a more comprehensive assessment by considering multiple emotions and indicating the confidence of each emotion's presence in a given face.}
  \label{fig:expression_circle}
\end{figure}

Further complicating the situation is the fact that due to the dynamic changes over the facial muscles \cite{cohn2004timing, ekman1997face, szajnberg2022face}, some facial expressions may not exclusively convey a single emotion, with individuals expressing mixed emotions in different emotional states—referred to by some researchers as \textit{compound expressions} \cite{du2015compound, li2017reliable, fabian2016emotionet, kollias2023multi, liu2022mafw, du2014compound, guo2018dominant}. Fig. \ref{fig:expression_circle} illustrates the combination of two expressions in a single facial image. In this figure, Happy-Contempt, Disgust-Anger, Sad-Neutral, and Fear-Surprise are jointly mixed in a facial image. Cultural differences represent another significant factor impacting emotional facial expressions and their perception, particularly among individuals from diverse cultural backgrounds~\cite{ekman1987universals, matsumoto2001handbook, matsumoto2019handbook, keltner2017understanding}. Additionally, expressing facial expressions is a dynamic, time-varying behavior, and in wild facial datasets, we capture only a snapshot of a person's evolving expression of emotion in still images. Consequently, it becomes challenging for humans to consistently and accurately judge the facial expressions. Hence, human annotators may not unanimously agree on emotion labels of others in complex, real-world environments. As a result, labels assigned to facial images in well-known datasets collected in wild settings, such as AffectNet \cite{mollahosseini2017affectnet}, RAF-DB \cite{li2017reliable}, and FER2013 \cite{goodfellow2013challenges}, are often noisy and unreliable.


As mentioned above, there is often disagreement between humans when annotating facial expressions, explicitly affecting the existing FER datasets~\cite{mollahosseini2017affectnet, li2017reliable, goodfellow2013challenges} and, ultimately, the automated FER models. While crowd-sourcing~\cite{li2017reliable} (using multiple trained human annotators) can alleviate this issue, the reported agreement between annotators is usually less than 68\%~\cite{mollahosseini2017affectnet}. This issue stems from the fact that assigning a single label (emotion) to an image might not be the right approach for annotating expressions, as some facial images express compound emotions. To address this concern, we propose an alternative approach where an image is annotated with more than one emotion label (which we refer to as \textit{soft-labels}), each with different degrees of confidence.

AffectNet~\cite{mollahosseini2017affectnet} is the largest publicly available in-the-wild facial expression dataset, containing both categorical~\cite{ekman1971constants} and dimensional (valence and arousal~\cite{russell1980circumplex}) labels. Despite its extensive use and application by researchers, AffectNet has several shortcomings and limitations that require further consideration. Firstly, although 450K out of one million images in AffectNet are annotated by human experts, the labels are noisy. In fact, each image is labeled only by one annotator, significantly detracting from the reliability of the labels. Hence, the potential noisy labels in AffectNet may have adversely contributed to the accuracy of FER models trained on AffectNet thus far. Secondly, only one label per image is given to AffectNet images, and as discussed before \cite{mollahosseini2017affectnet}, the dataset is collected by crawling the web, often producing images that contain compound emotions. Furthermore, the metadata (such as facial landmark points) released with AffectNet is noisy, as the algorithm used to extract facial landmark points has significantly improved in recent years. Additionally, the dataset lacks other metadata such as age, race, gender, and head pose, which are crucial in various affective computing applications.

To address these issues, this paper introduces AffectNet+, a revised version of AffectNet, which will be publicly available to the research community\footnote{A copy of AffectNet+ will be available for the interested researchers via: \href{http://mohammadmahoor.com/databases-codes/}{http://mohammadmahoor.com/databases-codes/}}. Although the concept of \textit{soft-label} is used in affective computing, there is no dataset covering this feature. AffectNet+ provides a novel approach to facial expression datasets, termed \textit{soft-labeling}. In contrast to the traditional method of assigning a single \textit{hard-label} to a facial image, \textit{soft-labeling} involves allocating multiple labels with varying degrees of confidence. In other words, a probability score is assigned to each of the seven emotion labels (plus an additional score for a Neutral label) that may be perceived when observing an image. Following this approach, we provide a new annotation vector named \textit{soft-label}, containing eight \textit{independent} probability scores corresponding to each emotion for every facial image in AffectNet+. Fig.~\ref{fig:expression_circle} illustrates examples of facial expressions with \textit{soft-labels}, where an image conveys two emotions with a high probability. Moreover, AffectNet+ categorizes the AffectNet images into three exclusive subsets based on the difficulty of recognizing facial expressions. These categories, denoted as \textit{Easy}, \textit{Challenging}, and \textit{Difficult}, are applied to both the training and validation sets.

\begin{table*}[t] 
\caption{Distribution of multi-annotated-set (MAS) and public AffectNet~\cite{mollahosseini2017affectnet} dataset. MAS is a private set of images labeled with at least two annotators.}
\label{tbl:expression_distribution}
\centering
\small
\resizebox{0.9\textwidth}{!}
{{
\begin{tabular}{lccccccccccc}
\hline
                           &            & Overal & Neutral & Happy  & Sad   & Surprise & Fear & Disgust & Anger & Contempt & Other  \\ \hline
\multirow{2}{*}{MAS}  & Train (train-MAS)      & 35250  & 4802    & 11183  & 3428  & 1598     & 1213 & 861     & 2648  & 785      & 8732   \\
                           & Validation (test-MAS) & 800    & 100     & 100    & 100   & 100      & 100  & 100     & 100   & 100      & 0      \\ \hdashline
\multirow{2}{*}{Public set} & Train      & 456349 & 80276   & 146198 & 29487 & 16288    & 8191 & 5264    & 28130 & 5135     & 137380 \\
                           & Validation & 5500   & 500     & 500    & 500   & 500      & 500  & 500     & 500   & 500      & 1500      \\ \hline
\end{tabular}
}}
\end{table*}

To create the \textit{soft-labels} for AffectNet+, we utilize a subset of AffectNet dataset containing 36K facial images, annotated by at least two human annotators. This subset provides more reliable labels compared to the single-annotator AffectNet training and validation sets. This subset is referred to as \textit{multi-annotated-set} (MAS). Table~\ref{tbl:expression_distribution} describes the MAS and AffectNet dataset. We propose two methods for creating \textit{soft-labels}: \textbf{\textit{1- Ensemble of binary classifiers}}, and \textbf{\textit{2- Action unit (AU)-based classifier}}. The "ensemble of binary classifiers" approach consists of training a set of binary classifiers (see Fig.~\ref{fig:arch_ens}), each designed to predict the probability score of a specific facial expression given an image (\textit{e.g.,} a model predicting the probability score of Happy versus all other facial expressions). The "AU-based classifier" leverages the overlap of AUs associated with facial expressions, defined by the Emotional Facial Action Coding System (EMFACS) \cite{ekman1978facial}. Specifically, for each emotion class, we train a binary classifier to jointly learn an AU-based representation vector as well as a binary class label (see Fig.~\ref{fig:arch_au_models}).

By calculating the probability vectors of the aforementioned classifiers, we designate a \textit{soft-label} vector to each image. Then, we compare the achieved \textit{soft-label} vector with the class label assigned by the annotator and categorize all the images in the AffectNet datset into Easy, Challenging, or Difficult subsets.

The contributions of our approach are summarized as follows:
\begin{itemize}
\item We introduce the notion of \textit{soft-labels} for facial expressions datasets, which could provide more realistic description of facial expressions.
\item We propose an automatic method to sub-categorize AffectNet into three subsets based on the level of difficulty of recognizing expressions in each image.
\item We introduce AffectNet+, the next-generation of facial expression dataset, which contains \textit{soft-labels} and other metadata, including age, gender, ethnicity, valence, arousal, head pose, and facial landmark points.
\end{itemize}

In the remainder of this paper Sec.~\ref{SEC:RELATED_WORK} reviews the related works. Sec.~\ref{SEC:METHODOLOGY} describes the proposed methodology for creating \textit{soft-labels}. Sec.~\ref{SEC:EXPERIMENTAL_RESULTS} discusses the experimental results. Sec.~\ref{SEC:SUBJECTIVE_EVALUATION} demonstrates the subjective evaluation of \textit{soft-label}. Sec. \ref{SEC:FUTURE_RESEARCH} highlights the open problems in FER using AffectNet+. Finally, Sec.~\ref{SEC:CONCLUSION} concludes the paper with some discussions on the proposed method.


\section{Related Works}\label{SEC:RELATED_WORK}
In this section, we review the major studies on the AffectNet database problems, as well as the researches focused on the compound datasets and \textit{soft-labeling} concepts in FER.
\subsection{FER Using AffectNet}
\label{fer_using_affectnet}
The AffectNet database is the largest in-the-wild dataset in existence today, includes 1 million images. In reviewing the SOTA papers that used AffectNet, we recognized three main challenges that researchers mainly dealt with \textbf{1)} uncertainty in the emotion labels, \textbf{2)} imbalanced data, and \textbf{3)} lack of data diversity. These challenges stem from the nature of the data distribution and the nature of the images posted on the web as the main source used to collect the images and create AffectNet. In the following, we explain these difficulties and some of the provided solutions. 

\textbf{Uncertainty in emotion labels:} Uncertainty in FER occurs when it is difficult for annotator (human or model) to determine the precise expression for a given facial image (see Fig.~\ref{fig:expression_circle}). Deep metric learning-based methods~\cite{fard2022ad, farzaneh2021facial}, self-learning~\cite{han2019deep}, latent space analysis~\cite{she2021dive}, and label-smoothing~\cite{zhang2021delving} are the most notable approaches proposed to deal with label uncertainty.

To handle the problem originated by label uncertainty, Gera~\etal~\cite{gera2022cern} utilized a lightweight network structure to combine the attention area with the local-global features to alleviate the noisy data. By considering the overlap between the expressions, Lang~\etal~\cite{lang2022multi} offered a three-step deep learning approach to group similar features, extract intra-class distribution, and finally distinguish similar expressions. Other approaches took advantage of AUs to deal with label uncertainty~\cite{liu2022uncertain, savchenko2022video}. Liu~\etal~\cite{liu2022uncertain} used AUs to find the most reliable image data, while Savchenko~\cite{savchenko2022video} combined AUs with valence-arousal to deal with noisy samples. Hasani~\etal~\cite{hasani2020breg} changed the shortcut passing method of the ResNet~\cite{he2016deep} model to a trainable transformer, to extract less correlated features. The relation between the accuracy of the model and the data distribution was studied by Dominguez-Caten~\etal~\cite{dominguez2022assessing}. They concluded that the balance between other facial attributes, such as gender and race, can improve the accuracy of the model. Another study by Su~\etal~\cite{su2022using}, as well as Heidari and Iosifidis~\cite{heidari2022learning}, showed the importance of compositional information between adjacent pixels in extracting robust features. Inspired by control theory, Wang~\etal~\cite{wang2022bias} developed transmitters for making a feedback cycle between regular one-hot label predictors and probabilistic label predictors, to generate \textit{soft-labels} for the images. To cope with label uncertainty, \textit{soft-labeling} was studied by Zhang~\etal~\cite{zhang2022man}. 

\textbf{Imbalanced data distribution:} This problem in FER originates from inequality between the number of samples per class. Table~\ref{tbl:expression_distribution} illustrates the distribution of various emotions in AffectNet. As this table shows, AffectNet is an imbalanced database. For instance, 32\% of the images in AffectNet are labeled Happy, while only 2\% of them are labeled Fear.  Data manipulation and model generalization~\cite{gao2023ssa, ma2022relation, zeng2022face2exp, heidari2022learning, jiang2021boosting} are among the most common approaches to tackle imbalanced data in AffectNet.

\textbf{A)} Data manipulation refers to up-sampling, down-sampling, and data knowledge sharing. For instance, Gao~\etal~\cite{gao2023ssa} extracted a subcategory for each expression before feeding their neural network. Lang~\etal~\cite{lang2022multi} considered only a third of the whole training set in the AffectNet dataset. In contrast, Gera~\etal~\cite{gera2022cern} upsampled the data through regular augmentation methods. Some research leveraged unsupervised and semi-supervised data to solve imbalanced data problems. While Jiang~\etal~\cite{jiang2021boosting} approached the imbalance data using semi-supervised learning, Zeng~\etal~\cite{zeng2022face2exp} combined unsupervised face recognition data with supervised AffectNet images to make a feedback-based adaptive network.

\textbf{B)} Model generalization approaches focus mainly on the objective functions to minimize the prediction error. Gong~\etal~\cite{gong2022effective} combined Focal Smoothing (FS) and Aggregation-Separation (AS) loss functions as EAFR loss. Similar study, by Li~\etal~\cite{li2019separate}, proposed a loss function for extracting basic facial expressions. Another method for confronting the imbalanced data was the weighted regularization method~\cite{liu2022uncertain}. Ma~\etal~\cite{ma2022relation} designed a cascade feature-augmentation method to preserve geometrical features and improve model generality by maximizing intra-sample and minimizing inter-sample similarities.

\textbf{Lack of data diversity:} This challenge in FER refers to the unevenness of demographic factors in a dataset, such as race, age, and gender, as well as some extrinsic factors, like head pose, occlusion, and illumination. This bias is problematic even in the AffectNet dataset despite its very large size. For instance, the number of images of males is nearly double that of images of females. We reported the data distribution over all the demographic factors in the Supplementary Materials. Researchers offered approaches to address this problem, such as focusing on regions of interest, ensemble learning, and domain adaptation.

\textbf{A)} Focusing on the regions of interest, i.e., exploring the most relevant parts of the facial image, is a solution to cope with the lack of data diversity. Zhang and Yu~\cite{zhang2022improving} turned to find a unique pattern map that transfers all the data of a specific class to a single pattern, different from the other classes. Another study considered the attention area problem as a multi-dimensional issue~\cite{gao2023ssa}. They combined spatial and spectral information and then extracted the relation between the AUs. Landmark detection and pyramid image scaling were other approaches for concentrating on the attention area~\cite{liu2023joint, zheng2022poster, kolahdouzi2022facetoponet}. Zheng ~\etal~\cite{zheng2022poster} suggested a cross-fusion transformer to take advantage of the landmarks to force the model to focus on the most related areas. On the other hand, Liu~\etal~\cite{liu2023joint} created a hierarchical attention map, where they cropped the attention area and skipped the rest of the image. 

\textbf{B)} Recent ensemble learning methods mainly provide parallel convolutional neural networks to extract robust features to address the lack of diversity. To alleviate this problem, Zia~\etal~\cite{ullah2022emotion} combined the features extracted by three VGG-19~\cite{simonyan2014very}, Inceptuion-V3~\cite{szegedy2016rethinking}, and ResNet-50~\cite{he2016deep} models to make a majority voting decision over the expressions. OANet~\cite{wang2021oaenet} was an oriented attention network structure that utilizes different networks in parallel and series, for diverse feature extraction and expression recognition.

\textbf{C)} Vision transformers were another approach to tackle the lack of diversity in AffectNet dataset. TransFER~\cite{xue2021transfer} model explored the relationship between different facial features. Dresvyanskiy~\etal~\cite{dresvyanskiy2022end} used an LSTM-RNN model alongside two different modalities of audio and video to transfer and fuse their knowledge. Rescigno~\etal~\cite{rescigno2020personalized} presented a combination of valence-arousal and facial features to exploit more robust features. Schiller ~\etal~\cite{schiller2020relevance} utilized an encoder-decoder to extract the saliencies on the expression and then fed the masked version of the input samples to the model to mitigate the lack of diversity.

Although the aforementioned methods mitigate the AffectNet dataset limitations, they are not a certain solution for the AffectNet complexity. How can we look at the data more realistically? Are the facial expressions explicitly separable? Are the facial AUs unique for any facial expression? What if we rethink the facial expressions in a way that any facial image can convey a portion of multiple expressions, simultaneously? The solution to these questions could be find in compund labeling and \textit{soft-labeling}.

\subsection{Compound FER Datasets and Soft-Labeling}
\label{compound_fer_datasets_and_soft_labeling}
Most facial expression recognition datasets are annotated with six basic facial expression labels and Neutral~\cite{lucey2010extended, mollahosseini2016facial, dhall2012collecting, goodfellow2013challenges, BarsoumICMI2016, zhang2018facial}. However, in some datasets, Contempt is added as the seventh basic expression~\cite{mollahosseini2017affectnet}. It is argued that sometimes these expressions are not explicitly separable (i.e., the uncertainty problem, discussed in Section \ref{fer_using_affectnet}). In other words, there are many cases in which more than one expression is included in a facial image. Some researchers created \textit{compound datasets} to deal with this problem. \textbf{They included at most two expression labels for the images, but without considering the intensity of each expression.} On the other hand, some researchers have worked on the idea of \textit{soft-labeling}, where they calculate the intensity of expressions in each facial image, but apply just one label to the facial image. However, to the best of our knowledge, no dataset exists with multiple labels with different intensities assigned to the facial images.

\subsubsection{Compound Datasets} 
RAF-DB~\cite{li2017reliable} is a manually annotated dataset, including six basic expressions, accompanied by twelve compound expressions, such as Happily-Surprised and Fearfully-Disgusted. FER+~\cite{barsoum2016training} is the new version of FER-2013~\cite{goodfellow2013challenges} dataset. This dataset includes eight expressions in the form of single and compund label expressions. EmotioNet~\cite{fabian2016emotionet} is another FER in-the-wild dataset with compound labels. They considered 23 basic expressions as descriptors of the dataset, where fourteen of them were compound (pair) expressions. C-EXPR-DB~\cite{kollias2023multi} is a manually annotated in-the-wild dataset, annotated by 12 compound expressions, including 400 videos (200K frames). In 2022, Liu~\etal~\cite{liu2022mafw} released the MAFW compound multi-modal dataset, containing more than 10K video clips, accompanied by audio and text descriptors. Barsoum~\etal\cite{BarsoumICMI2016} worked on the dataset FER-2013~\cite{goodfellow2013challenges} and re-labeled this dataset in a compound labeling format.

In addition to these in-the-wild datasets, there are two compound lab-controlled datasets. Du~\etal~\cite{du2014compound} created a dataset, including 21 compound expressions of 230 subjects. This dataset includes the expressions and the intensity of the AUs. As the second compound dataset, iCV-MEFED~\cite{guo2018dominant}, containing 31250 facial images, targeted 125 subjects in a controlled environment and assigned 49 compound expressions to the facial images (plus Neutral). For any subject, they defined 50 compound expressions and captured 5 images per person-expression. 

The aforementioned datasets highlight the essence of paying more attention to the compound expressions in FER. However, it is notable that all the reviewed datasets provide neither more than one combination of the labels, nor the intensity of each expression. For more information about the datasets in FER, we refer our readers to the Supplementary Materials.

\subsubsection{Soft-Labeling} 
Some research in expression recognition has recently focused on extracting \textit{soft-labels} rather than \textit{hard-labels}~\cite{wang2022bias, ming2022soft, jiang2023joint, ma2023transformer, gan2019facial, liu2021facial, lukov2022teaching}.
Gan~\etal~\cite{gan2019facial} proposed a model to discover the co-occurrence of multiple expressions in a single image. They initially trained a model to generate a probability vector over the expressions. These probabilities were then perturbed to generate \textit{soft-labels}. In the last step, the \textit{soft-labels} were used in another model to find the intrinsic relation between the expressions in an image. In another line of research, Liu~\etal~\cite{liu2021facial} studied non-verbal behavior in schools, using infrared images. They initially extracted the similarity between different expressions and then fed the data into their CDLLNet model to learn the Cauchy distribution over the expressions. This method enabled them to have multiple expressions with different intensities for a single image. To relax the effect of noisy samples, Lukov~\etal~\cite{lukov2022teaching} developed a Soft Label Smoothing (SLS) model to smooth the logits. In this model, instead of labeling the facial expressions, a probability vector was generated to show the correlation of the expressions in an image. 

All these models worked on \textit{soft-labeling}, but they generated their \textit{soft-labels} with different methods and had no evaluation set to evaluate or compare their approach. Therefore, having a dataset including \textit{soft-labels} could provide more general and robust models. \textit{Soft-labeling} methods and the aforementioned compound FER datasets highlight the necessity of paying attention to the \textit{soft-labeled} facial expression recognition datasets. To cover this essence, this paper introduces the AffectNet+ dataset, including \textit{soft-labels}, three categorizations of the data, and some useful metadata, that could open new perspectives toward FER studies.


%
%

%
\begin{figure*}[t]
  \centering
  \includegraphics[width=0.9\textwidth]{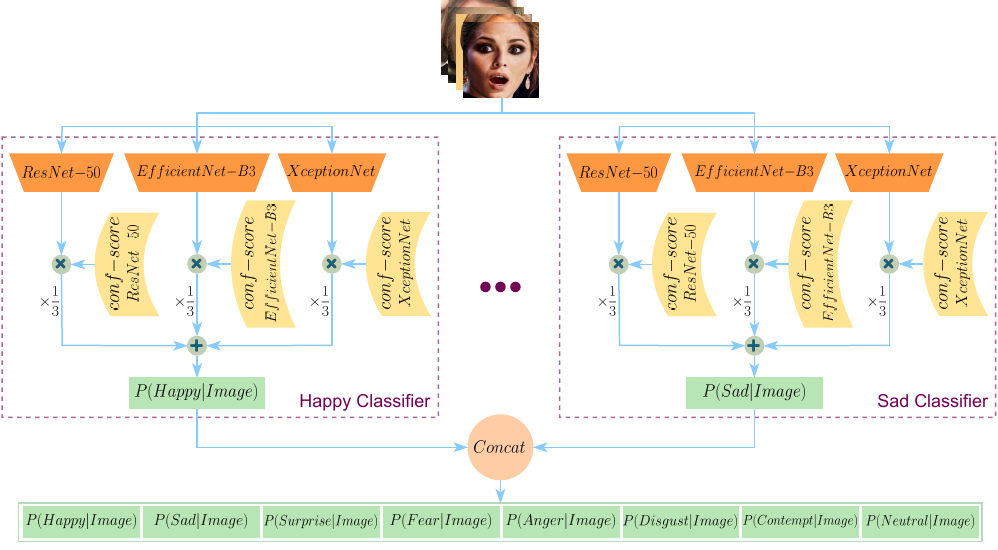}
  \caption{Architecture of ensemble of binary classifiers (EBC model), as the initial step of the \textit{soft-labeling} process. It contains ensemble of three ResNet-50~\cite{he2016deep}, EfficientNet-B3~\cite{tan2019efficientnet}, and XceptionNet~\cite{chollet2017xception} classifiers, for any expression. There are eight instances of this network architecture, trained for each expression in a binary one-vs-rest method. Finally, their output aggregates to make the expression vector.}
  \label{fig:arch_ens}
\end{figure*}

\section{Methodology} \label{SEC:METHODOLOGY}
In this section, we first explain our novel Soft-FER and the process of creating \textit{soft-labels}. Afterward, we introduce the AffectNet+ database and its Easy, Challenging, and Difficult subsets. Finally, we explain the metadata we updated or added to AffectNet+.

\begin{table}[b] 
\caption{Action units for different expressions \cite{ekman1978facial, martinez2017automatic}. Different subsets of the corresponding AUs will create an expression. For instance, AU1, AU4, AU15, AU17 create Sad expression, while another combination could be AU1, AU4, AU6, AU11, AU15.}
\label{tbl:action_units_for_exp}
\centering
\small
\resizebox{0.34\textwidth}{!}
{{
\begin{tabular}{lccccccc}
\hline
\multicolumn{1}{l}{}           & \multicolumn{7}{c}{Action Units}                                                                                                                                                                              \\ \hline
\multicolumn{1}{l}{Happy}      & \multicolumn{7}{l}{6, 12, 25}                                                                                                                                                                                 \\ 
\multicolumn{1}{l}{Sad}        & \multicolumn{7}{l}{1, 4, 6, 11, 15, 17}                                                                                                                                                                       \\ 
\multicolumn{1}{l}{Surprise}   & \multicolumn{7}{l}{1, 2, 5, 26, 27}                                                                                                                                                                           \\
\multicolumn{1}{l}{Fear}       & \multicolumn{7}{l}{1, 2, 4, 5, 20, 25, 26, 27}                                                                                                                                                                 \\
\multicolumn{1}{l}{Anger}      & \multicolumn{7}{l}{4, 5, 7, 10, 17, 22, 23, 24, 25, 26}                                                                                                                                                       \\ 
\multicolumn{1}{l}{Disgust}    & \multicolumn{7}{l}{9, 10, 16, 17, 25, 27}                                                                                                                                                                      \\ 
\multicolumn{1}{l}{Contempt}   & \multicolumn{7}{l}{12, 14}    \\                                      \hline
\end{tabular}
}}
\end{table}

\subsection{Soft-FER} \label{SEC:METHODOLOGY_SFER}
Facial expressions are the result of facial muscle movements, which can be coded in terms of action units (AUs). EMFACS~\cite{ekman1978facial} describes many combinations of facial muscle movements related to each expression. According to EMFACS, for almost all basic facial expressions, there exists more than one combination of AUs. For instance, Happy expression can be shown by the activation of specific AUs, such as AU6 and AU12, or solely AU12. These combinations illustrate the intra-class variation in FER. Likewise, EMFACS shows a high correlation in AUs for specific emotions. For example, action units 6, 12, and 25 correspond to Happy emotion, while AU12 and AU14 correspond to Contempt.  This correlation between the action units highlights inter-class similarities in FER. Tables~\ref{tbl:action_units_for_exp} and~\ref{tbl:action_units_correlation} show the AUs for each emotion class and the correlation between them, respectively.

Hence, people should potentially perceive more than one specific facial expression from a facial image in many cases. In fact, by assigning \textit{only one} emotional label to a facial image we are ignoring the valuable information that can be utilized to provide a more comprehensive explanation of facial expression. We argued that widely used Hard-FER, where we assign one label to an image, needs further consideration, and accordingly, we proposed Soft-FER as a solution. In our proposed Soft-FER, we measured the probability score of the existence of all the facial expressions for each image as follows in Eq.~\ref{eq:SL_single}:
\begin{align}\label{eq:SL_single}
P(emo_i| img_k) ~ \qquad \forall ~ emo_i \in \text{EMOTIONS},
\end{align}
where EMOTIONS = \{Neutral, Happy, Sad, Surprise, Fear, Disgust, Anger, and Contempt\},  $i \in \{0, 1, .., 7\}$ indicating the $i^{th}$ expression, $k \in \{0, 1, ..., N\}$, and $N$ is the number of images in the dataset. We used a neural network to estimate the corresponding probability $P$. Using Eq.~\ref{eq:SL_single}, we defined a \textit{soft-label} vector, $SL_k$, corresponding to $img_k$ as follow in Eq.~\ref{eq:SL}:
\begin{align}\label{eq:SL}
 \begin{matrix*}[c]
        P_{ik} := P(emo_i| img_k), \\ \\
        SL_k := \{P_{0k}, P_{1k}, ..., P_{7k}\}.
    \end{matrix*}
\end{align}

\begin{table}[b] 
\caption{Correlation between the action units, regarding each emotion class. Each value shows the number of common action units between two expressions, using EMFACS~\cite{ekman1978facial}.}
\label{tbl:action_units_correlation}
\centering
\small
\resizebox{0.5\textwidth}{!}
{{
\begin{tabular}{lccccccc}
\hline 
\multicolumn{1}{c}{}         & \multicolumn{1}{c}{Happy} & \multicolumn{1}{c}{Sad} & \multicolumn{1}{c}{Surprise} & \multicolumn{1}{c}{Fear} & \multicolumn{1}{c}{Disgust} & \multicolumn{1}{c}{Anger} & \multicolumn{1}{c}{Contempt} \\ \hline

Happy   & \multicolumn{1}{c}{-}     & \multicolumn{1}{c}{1}   & \multicolumn{1}{c}{0}        & \multicolumn{1}{c}{1}    & \multicolumn{1}{c}{1}     & \multicolumn{1}{c}{1}       & \multicolumn{1}{c}{1}        \\ 

Sad   & \multicolumn{1}{c}{1}     & \multicolumn{1}{c}{-}   & \multicolumn{1}{c}{1}        & \multicolumn{1}{c}{2}    & \multicolumn{1}{c}{1}     & \multicolumn{1}{c}{2}       & \multicolumn{1}{c}{0}        \\ 
Surprise & \multicolumn{1}{c}{0}     & \multicolumn{1}{c}{1}   & \multicolumn{1}{c}{-}        & \multicolumn{1}{c}{5}    & \multicolumn{1}{c}{1}     & \multicolumn{1}{c}{2}       & \multicolumn{1}{c}{0}        \\ 
Fear     & \multicolumn{1}{c}{1}     & \multicolumn{1}{c}{2}   & \multicolumn{1}{c}{5}        & \multicolumn{1}{c}{-}    & \multicolumn{1}{c}{2}     & \multicolumn{1}{c}{4}       & \multicolumn{1}{c}{0}        \\ 
Disgust  & \multicolumn{1}{c}{1}     & \multicolumn{1}{c}{1}   & \multicolumn{1}{c}{1}        & \multicolumn{1}{c}{2}    & \multicolumn{1}{c}{-}     & \multicolumn{1}{c}{4}       & \multicolumn{1}{c}{0}        \\
Anger    & \multicolumn{1}{c}{1}     & \multicolumn{1}{c}{2}   & \multicolumn{1}{c}{2}        & \multicolumn{1}{c}{4}    & \multicolumn{1}{c}{4}     & \multicolumn{1}{c}{-}       & \multicolumn{1}{c}{0}        \\ 
Contempt & \multicolumn{1}{c}{1}     & \multicolumn{1}{c}{0}   & \multicolumn{1}{c}{0}        & \multicolumn{1}{c}{0}    & \multicolumn{1}{c}{0}     & \multicolumn{1}{c}{0}       & \multicolumn{1}{c}{-}        \\ \hline
\end{tabular}
}}
\end{table}

As Fig~\ref{fig:expression_circle} shows, \textit{soft-labels} are more explanatory compared to \textit{hard-labels} as they explicitly present the similarity between a facial image $img_k$ and all the emotions in $EMOTIONS$ set. In fact, hard-label does not consider the variation within an emotion class. For example, very happy versus slightly happy can be potentially confused with the Neutral expression. It also distorts the similarity between different emotion classes. It means that a facial image can be perceived as both Anger and Fear, as there is a high correlation between the AUs corresponding to such emotions. On the contrary, \textit{soft-labels} do not have these drawbacks as it considers the probability score of the existence of all the emotion classes for a facial image. Consequently, the machine learning model would learn the different variations of a specific emotion class, as well as the similarities between different classes. 

To the best of our knowledge, there exists no FER dataset providing \textit{soft-labels}. Creating such labels necessitates training annotators in accordance with Soft-FER methodology, which demands a significant investment of both time and financial resources. Hence, in AffectNet+ we attempt to automatically generate \textit{soft-labels} using deep learning-based methods. 

In order to generate \textit{soft-labels} automatically for both the training and validation sets of AffectNet, we used our multi-annotated set (MAS). For more detail on the MAS refer to Supplementary Materials. We divided MAS into training and test sets. For each emotion, we selected 100 images with the most obvious facial expression as the test set, called test-MAS. To clarify, if all the human annotators agreed on a facial expression the respective image was a candidate for our test set. The rest of the images in MAS were considered as the training set, which we refer to as train-MAS. Table~\ref{tbl:expression_distribution} shows the training and test set configuration created from the multi-annotated set (MAS). 

In the next step, we designed and utilized two solutions to calculate  \textit{soft-labels} for each image in the training and validation set of the AfectNet dataset. particularly, this paper introduces AffectNet+ by adding \textit{soft-labels}, three level of data complexity, as well as a set of additional metadata, to the AffectNet dataset. To assign \textit{soft-label} to each image, we calculated the probability score of all the emotions. Accordingly, we proposed the following methods: \textbf{1- Ensemble of binary classifiers}, and \textbf{2- AU-based classifier}. In the following, we explain each method.

\subsection{Ensemble of Binary Classifiers (EBC)} \label{SEC:METHODOLOGY_ens_cls}
Categorical state-of-the-art models~\cite{fard2022ad, hasani2020breg, farzaneh2021facial} face a high confusion rate while distinguishing between emotions that exhibit significant similarities, such as Neutral and Contempt. To alleviate this challenge, we proposed 8 binary classifiers, each trained to detect one facial expression in a one-vs-rest way. In fact, instead of using a convolutional neural network to predict the probability score of all the facial expressions at once, we introduced 8 different CNNs, each trained to detect only one facial expression. Moreover, to increase the confidence of the prediction, we utilized an ensemble of binary classifiers by the following CNNs: ResNet-50~\cite{he2016deep}, EfficientNet-B3~\cite{tan2019efficientnet}, and XceptionNet~\cite{chollet2017xception}. Fig. \ref{fig:arch_ens} demonstrates the architecture of our binary classifier. We ensembled three of these binary classifiers, with different network architectures, to achieve more robust results.

\textbf{Training:} \label{SEC:METHODOLOGY_ens_cls_train}
Training binary classifiers using train-MAS needed first choosing a set of positive and negative samples. Assume we train a binary classifier to predict Sad emotion, all the images in the training set annotated as Sad are taken as the positive samples, and the rest can be chosen as the negative samples. One naive approach is to choose all the images labeled as desired facial expressions as positive and the rest as negative samples, resulting in an imbalanced training set, and accordingly a biased classifier. Thus, we proposed a novel positive-negative selection strategy to ensure the high accuracy of the classifiers.

We utilized the correlation between the AUs corresponding to different emotions to choose the ratio of the negative samples. For training a binary classifier, to detect the facial expression of emotion $emo_i$, we selected the maximum number of negative samples from the images annotated as $emo_j$, where $emo_j$ has the highest AU correlation with $emo_i$. As certain emotions may not share any similar AUs, we always chose 20\% of the negative samples randomly to ensure a uniform distribution from all the other emotions. The remaining negative samples were allocated proportionally based on the similarity ratio of the corresponding AUs between $emo_i$ and the emotions that share similar AUs. Table~\ref{tbl:action_units_for_exp} shows the AUs associated with each emotion class. 

\textbf{Confidence Score Calculation:} \label{SEC:METHODOLOGY_ens_cls_conf_score}
We introduced the term \textit{confidence score} to indicate the level of trustworthiness in the prediction of each binary classifier model. For each binary classifier, the confidence score is defined as the average per-class accuracy. Since we followed a one-vs-rest training approach, the number of negative samples was far more than the number of positive samples. To tackle this imbalanced distribution, we defined the confidence score of each emotion class $emo_i$ as follows in Eq.~\ref{eq:SC_bin_cls}:
\begin{align}\label{eq:SC_bin_cls}
CS(emo_i):= \frac{1}{2}(\frac{TP_{{emo}_i}}{TP_{{emo}_i}+FP_{{emo}_i}} + \frac{TN_{{emo}_i}}{TN_{{emo}_i}+FN_{{emo}_i}}).
\end{align}
We used the confidence score for each binary classifier in the inference, for adjusting the probability score assigned to a facial image considering each emotion class. Table~\ref{tbl:conf_score} shows the confidence scores of each binary classifier. It is also notable that we report this score as the average accuracy, $\overline{Acc}$, in Sec.~\ref{SEC:EXPERIMENTAL_RESULTS}. 

\textbf{Inference:} \label{SEC:METHODOLOGY_ens_cls_inf}
We also leveraged the \textit{semantic score} associated with the ensemble of the binary classifiers, called $SC^{EB}$. This score indicates the existence of the emotion $emo_i \in EMOTIONS$ in an arbitrary facial image $img_k$. It is calculated using the multiplication of the corresponding probability ($P$) and the confidence score ($CS^{EB}$), as follows in Eq.~\ref{eq:sem_score_bin_cls_single}:

\begin{align}
\label{eq:sem_score_bin_cls_single}
SC^{EB}(emo_i, img_k):= CS^{EB}(emo_i) \times P(emo_i| img_k).
\end{align}
In the ensemble of binary classifiers model, for each emotion class, we have three $P$ functions, with their corresponding confidence scores. For an emotion class $emo_i$, we calculated the ensemble of the semantic scores as the average score of three binary classifiers as follows in Eq.~\ref{eq:sem_score_bin_cls_all}:
%

\begin{align}\label{eq:sem_score_bin_cls_all}
 \begin{matrix*}[c]
        SC^{EB}_{Mean}(emo_i, img_k) := \sum_{j \in \{ RN, EN, XN\} }^{} SCP_{j}, \\ \\
        SCP_{j} = \frac{1}{3} CS^{EB}_{j}(emo_i) P_{j}(emo_i| img_k).
    \end{matrix*}
\end{align}
In this equation, $RN$, $EN$, and $XN$ refer to ResNet-50~\cite{he2016deep}, EfficientNet-B3~\cite{tan2019efficientnet}, and XceptionNet~\cite{chollet2017xception}, respectively. Table~\ref{tbl:bin_eval_test_MAS} shows per-class confidence scores for each classifier (indicated as $\overline{Acc}$). We will use these scores in Sec.~\ref{SEC:Creating_Soft-Labels} to calculate soft-labels.

\subsection{Action Unit (AU)-Based Classifier} \label{SEC:METHODOLOGY_au_cls}
In this section for each emotion class $emo_i$ we trained a model to learn the corresponding AU-based representation vector. We proposed a novel algorithm that utilizes the representation vector (AU vector), generated by each model to estimate the probability of the corresponding facial expression. In contrast to the ensemble of binary classifiers, which utilized \textit{hard-labels} for training, the AU-based classifier used an AU-based representation of each emotion, resulting in a fine-grained analysis of facial expressions.

Unlike the previous studies~\cite{lucey2010extended, tan2022emotion}, where neural networks were trained to learn AUs specifically for FER or valence-arousal estimation, our novel method leveraged AUs \textit{only} as a more comprehensive representation. We proposed a deep neural network that tends to learn the AUs presented in a given image.

As Table~\ref{tbl:action_units_for_exp} shows, for each emotion class, there exists a set of AUs, which can be used as a representation vector. The AU-based representation vector can explicitly convey the inter-class similarity. Thus, training a CNN model, to learn and capture the unique  AU-based representation vector of each emotion class, can potentially assist the neural network to better learn facial expressions from facial images.

\textbf{Training:} \label{SEC:METHODOLOGY_au_cls_train}
To train the AU-based classifier, we first defined the representation vector for each emotion class. We used 21 different AUs to model 7 basic facial expressions (refer to Table~\ref{tbl:action_units_for_exp}). Hence, the length of the representation vector was 21. We showed the set of AUs as follows in Eq.~\ref{eq:au_seq}:
\begin{align}\label{eq:au_seq}
\begin{matrix*}[r]
        AU:= \{1, 2, 4, 5, 6, 7, 9, 10, 11, 12, 14, 15, \\
                 16, 17, 20, 22, 23, 24, 25, 26, 27\}.
\end{matrix*} 
\end{align}

For each emotion $emo_i \in EMOTIONS$, we first referred to Table~\ref{tbl:action_units_for_exp} to identify the corresponding set of action units, denoted as $AU\_set_i$. Then, we constructed an AU-based representation vector $AU_i$ for $emo_i$, where all indices were initialized to zero except for those corresponding to the action units in $AU\_set_i$, which were set to 1. This resulted in a sparse vector where the majority of values were zero, indicating the absence of the corresponding AUs, while the non-zero values (ones) indicated the presence of the specific AUs associated with $emo_i$.

For each emotion $emo_i \in EMOTIONS$, we trained a model to learn the corresponding AU-based representation vector $AU_i$. As the synergy between two related tasks can improve the overall performance of the models~\cite{fard2021asmnet}, we designed our models to simultaneously generate the representation vectors, as well as performing a binary classification task.

\begin{figure*}[t]
  \centering
  \includegraphics[width=0.9\textwidth]{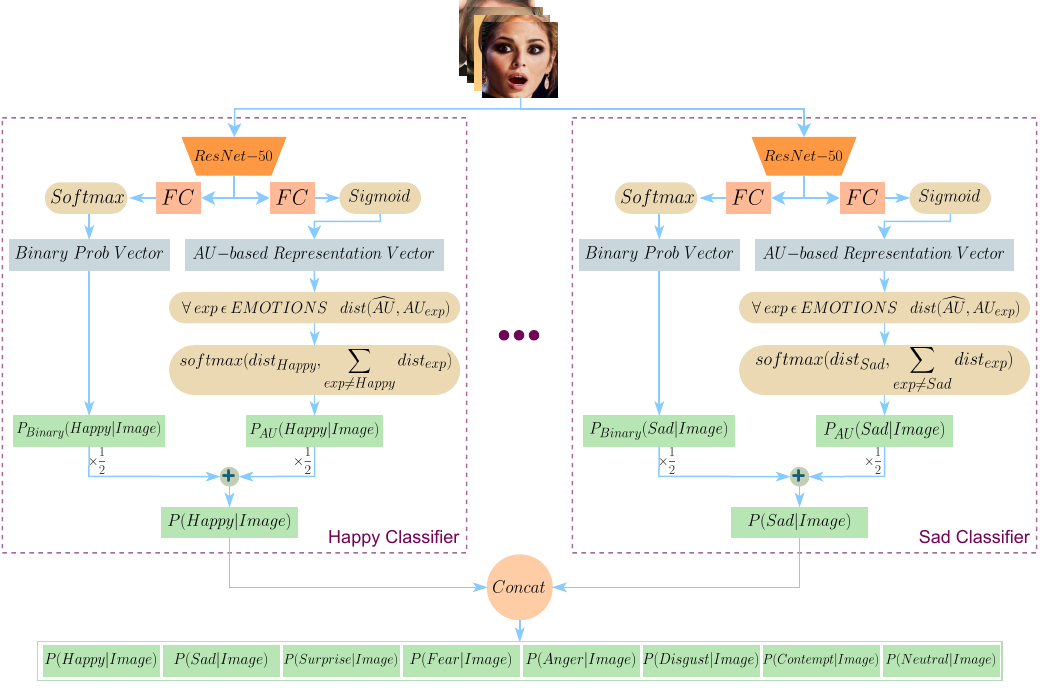}
  \caption{Architecture of the AU-based classifier for each expression, as the second model of the \textit{soft-labeling} process. For each emotion class, a multi-head ResNet-50~\cite{he2016deep} classifier is trained to simultaneously learn the features in the AUs and the expressions. Each model is trained to find the relation between the expressions and AUs. There are eight instances of this network architecture, trained for each expression. Similar to the initial model (EBC), each expression is trained in a binary one-vs-rest way, and their output aggregates to make the expression vector.}
  \label{fig:arch_au_models}
\end{figure*}

We followed the approach described in Sec.~\ref{SEC:METHODOLOGY_ens_cls_train} for choosing the positive and negative samples. As Fig.~\ref{fig:arch_au_models} illustrates, the multi-head model ResNet-50~\cite{he2016deep} that we used for our AU-based classifier, consisted of two fully connected (FC) layers. Each head was responsible for a specific task. A binary classifier head focused on labeling the input sample as a negative or positive sample. Another head extracted the AU-based representation vector. 

On the one hand, to train the binary classifier head, we used a Softmax activation function after the last fully connected (FC) layer, and binary cross entropy (CE) as the loss function. Thus, the output of the binary classification task was a \textit{2} dimensional vector called Binary Probability Vector (BPV).

On the other hand, to generate the AU-based representation vector, we utilized the Sigmoid activation function following the final FC layer. Using the Sigmoid function, we forced the model to learn the value of each element of the representation vector. Further, we used the multi-label cross entropy (CE) as the loss function. The output of this head was an AU-based representation vector with the size of \textit{21}. This vector later helped us to score each emotion based on its corresponding action units.

We created two one-hot weight maps, $\omega_{pos}$ and $\omega_{neg}$, where $\omega_{pos}$ showed the active AUs for an image, and $\omega_{neg}$ indicated its inactive AUs. The length of this positive and negative weight maps was equal to the length of the $AU_k$ (21). Finally, we defined our multi-label cross entropy loss as Eq.~\ref{eq:au_loss}:
\begin{align}\label{eq:au_loss}
\begin{split}
    & L^{Pos}_{k} :=  -\sum_{i=1}^{n=21} \omega_{pk}^{i} ~ AU^{i}_k ~ log(\hat{AU}^{i}_k), \\
    & L^{Neg}_{k} :=  -\sum_{i=1}^{n=21} \omega_{nk}^{i} ~ (1-AU^{i}_k) ~ log(1-\hat{AU}^{i}_k), \\
    & Loss := \sum_{k=1}^{N} L^{Pos}_{k} + L^{Neg}_{k},
\end{split}
\end{align}
where $AU_k$ and $\hat{AU}_k$ are the ground truth and the generated AU-based representation vectors, respectively, and $N$ is the number of training set samples. To explain more, we considered each element in $\hat{AU}_k$ as a binary classification task.

\textbf{Confidence Score Calculation:} \label{SEC:METHODOLOGY_au_cls_conf_score}
We introduced the confidence score as a metric to track the accuracy of the AU-based classifier. Since the AU-based classifier performs two tasks (binary classification, as well as generating an AU-based representation vector), we derived the prediction by taking the average of the probability scores associated with each task. Then we used the average accuracy as the confidence score. 

For the AU-based representation vector, we proposed a novel algorithm to assess the similarity between the predicted representation vector and the corresponding ground truth. We proposed a weighting strategy based on the ratio of the presence of an action unit in the emotion set $EMOTIONS$, and accordingly, assigned a score to each AU. We defined the score for each AU to be inversely proportional to the frequency of its presence within the emotion set. Hence, the less frequently an AU appears in the emotion set, the greater its score will be. To illustrate, AU14 exclusively appears in Contempt, while AU25 appears in four expressions, Happy, Fear, Disgust, and Anger. Hence, we assigned a score of 1 to the former action unit (AU 14) and $\frac{1}{4}$ to the latter (AU 25). In Eq.~\ref{eq:AUS}, we defined the score vector of the AUs, known as $AUS_{k_{1 \times n}}$, such that its $i^{th}$ element represents the score corresponding to the $i^{th}$ element of the AU.
\begin{align}\label{eq:AUS}
\begin{matrix*}[r]
        AUS:= \{ 0.33, 0.5, 0.33, 0.33, 0.5, 1.0, 1.0, 0.5, 1.0, 0.5, 1.0,\\
        1.0, 1.0, 0.33, 1.0, 1.0, 1.0, 1.0, 0.25, 0.25, 0.5 \}.
\end{matrix*} 
\end{align}

For an image $img_k$, we introduced the similarity vector $SV_{k_{1 \times 8}}$ in Eq.~\ref{eq:SV}, such that $j^{th}$ element represents the similarity between generated and ground truth AU-based representations. 
\begin{align}\label{eq:SV}
\begin{split}
& sim_{emo_j} := \sum_{i=0}^{n} AUS^{i} (AU^{i}_{emo_j} \hat{AU}^{i}),  \\
& SV_k := \{ sim_0, sim_1, ..., sim_7 \}. 
\end{split}
\end{align}
$AU_{emo_j}$ is the AU-based representation vector for the emotion class $emo_j \in EMOTIONS$, while $\hat{AU}$ shows the generated representation vector. $AU^{i}_{emo_j}$ and $\hat{AU}^{i}$ are the $i^{th}$ elements of $AU_{emo_j}$ and $\hat{AU}$, respectively. Likewise, $sim_{emo_j}$ is the weighted sum of non-zero elements in the generated $\hat{AU}$ and ground truth $AU$ of $emo_j$. It is notable that for Neutral, where all the elements in $AU_{Neutral}$ are zero, we define $sim_{Neutral}= 0.25$ as a hyper-parameter.

In the next step, we introduced the corresponding \textit{binary} similarity vector, $BSV_{k_{1 \times 2}}$,  as follows in Eq.~\ref{eq:BSV}:
\begin{align}\label{eq:BSV}
\begin{split}
BSV_k :=  \{ SV_k^{gt},~~~~~ \frac{1}{7}\times\sum_{i=0, i \neq gt }^{7}  SV_k^{i} \},
\end{split}
\end{align}
where $ gt \in \{0, ..., 7\}$ is the index of the ground truth emotion class. In fact, $BSV_k$ means that for the $img_k$, we calculated the score of the expected expression versus the average of the other expressions. Afterward, we calculated the AU-based \textit{binary} probability vector $APV_{k_{1 \times 2}}$ using the corresponding similarity vector $BSV_{k}$ as follows in Eq.~\ref{eq:PV}:
\begin{align}\label{eq:PV}
\begin{split}
APV_k := \{ \frac{   e^{BSV^{0}_k}  }{ \sum_{i=0}^{1} e^{BSV^{i}_k}} 
            ,~~~~~
            \frac{   e^{BSV^{1}_k}  }{ \sum_{i=0}^{1} e^{BSV^{i}_k}} 
\}.
\end{split}
\end{align}

In addition, for the binary classification task in Fig.~\ref{fig:arch_au_models}, we defined $BPV_{k_{1\times2}}$ as the binary probability vector associated with $img_k$. Finally, the element-wise sum between $BPV_k$ and $APV_k$ is used for the ultimate classification.
\begin{align}\label{eq:P}
P_k = \frac{1}{2} (BPV_k + APV_k).
\end{align}
We followed the approach described in Sec.~\ref{SEC:METHODOLOGY_ens_cls_conf_score}, and used the average accuracy as the confidence score. Table~\ref{tbl:conf_score} shows the confidence scores of each expression.

\begin{table*}[t!] 
\caption{Per-class confidence scores for EBC (ensemble of binary classifiers) and AU (AU-based classifier), in percent. The effect of the AU-based classifier on challenging expressions, like Sad, Fear, Contempt, and Surprise is inevitable.}
\label{tbl:conf_score}
\centering
\small
\resizebox{0.6\textwidth}{!}
{{
\begin{tabular}{lcccccccc}
\hline
& Neutral & Happy  & Sad    & Surprise & Fear   & Disgust & Anger  & Contempt \\ \hline
EBC   & 81.33    & 87.45   & 79.88   & 86.47     & 83.76   & 84.52    & 84.69   & 66.78  \\  
AU & 88.71    & 87.50   & 84.64   & 90.71     & 89.00   & 86.28    & 84.78   & 77.78    \\ \hline
\end{tabular}
}}
\end{table*}

\textbf{Inference:} \label{SEC:METHODOLOGY_au_cls_inf}
For any image in the training set of AffectNet, we measured the probability of the presence of the $emo_i \in EMOTIONS$, following the approach explained in Sec.~\ref{SEC:METHODOLOGY_au_cls}, using Eq.~\ref{eq:P}. We measured the AU-based Semantic Score ($SC^{AU}$) as follows in Eq.~\ref{eq:sem_score_AU}:
\begin{align}\label{eq:sem_score_AU}
SC^{AU}(emo_i, img_k):= CS^{AU}(emo_i) P(emo_i| img_k),
\end{align}
where $CS^{AU}$ is the confidence score of the AU-based classifier, calculated by Eq. \ref{eq:SC_bin_cls}, and $P$ is the AU-based binary probability vector introduced in Eq.~\ref{eq:P}. See Table~\ref{tbl:action_unit_eval_test_MAS} for per-class evaluation scores associated with the AU-based classifier.
\begin{table*}[t!] 
\caption{The distribution of AffectNet+ train set over different subsets of \textit{Easy}, \textit{Challenging}, and \textit{Difficult}. The \textit{Easy} subset determines the set of images that the model and the annotator agree on their expression. The \textit{Challenging} subset refers to the images that the annotator and the model do not agree on, but their label is in the model's top-3 predictions. The \textit{Difficult} subset determines the samples their label is out of the model's top-3 predictions.}
\label{tbl:subset_distribution_affectnet_plus}
\centering
\small
\resizebox{0.85\textwidth}{!}
{{

\begin{tabular}{lccccccccc}
\hline 
& Neutral  & Happy  & Sad   & Surprise & Fear & Disgust & Anger  & Contempt  & Overal                    \\ \hline
All         & 74,874  & 134,415   & 25,459  & 14,090   & 24,882   & 3,803  & 6,378   & 3,750  & 287,651   \\ \hdashline
Easy      
& \begin{tabular}[c]{@{}c@{}}   51,422      \\ (68.67\%)    \end{tabular}
& \begin{tabular}[c]{@{}c@{}}   115,934     \\ (86.25\%)    \end{tabular} 
& \begin{tabular}[c]{@{}c@{}}   8,171       \\ (32.04\%)    \end{tabular}  
& \begin{tabular}[c]{@{}c@{}}   4,914       \\ (34.87\%)    \end{tabular} 
& \begin{tabular}[c]{@{}c@{}}   10,651      \\ (42.08\%)    \end{tabular} 
& \begin{tabular}[c]{@{}c@{}}   987         \\ (25.95\%)    \end{tabular}   
& \begin{tabular}[c]{@{}c@{}}   1,698       \\ (26.62\%)    \end{tabular} 
& \begin{tabular}[c]{@{}c@{}}   477         \\ (12.72\%)    \end{tabular}
& \begin{tabular}[c]{@{}c@{}}   194,254         \\ (67.53\%)    \end{tabular}\\ \hdashline

Challenging 
& \begin{tabular}[c]{@{}c@{}}   14,669  \\ (19.59\%)\end{tabular} 
& \begin{tabular}[c]{@{}c@{}}   11,835  \\ (8.80\%)\end{tabular}  
& \begin{tabular}[c]{@{}c@{}}   11,067  \\ (43.46\%)\end{tabular} 
& \begin{tabular}[c]{@{}c@{}}   4,646   \\ (32.97\%)\end{tabular} 
& \begin{tabular}[c]{@{}c@{}}   8,837   \\ (35.51\%)\end{tabular} 
& \begin{tabular}[c]{@{}c@{}}   1,663   \\ (43.72\%)\end{tabular} 
& \begin{tabular}[c]{@{}c@{}}   2,270   \\ (35.91\%)\end{tabular} 
& \begin{tabular}[c]{@{}c@{}}   2,440   \\ (65.06\%)\end{tabular} 
& \begin{tabular}[c]{@{}c@{}}   57,427   \\ (19.96\%)\end{tabular} \\ \hdashline
Difficult        
& \begin{tabular}[c]{@{}c@{}}   8,783   \\ (11.73   \%)\end{tabular}  
& \begin{tabular}[c]{@{}c@{}}   6,646   \\ (4.94    \%)\end{tabular}    
& \begin{tabular}[c]{@{}c@{}}   6,221   \\ (24.43   \%)\end{tabular} 
& \begin{tabular}[c]{@{}c@{}}   4,530   \\ (32.15   \%)\end{tabular} 
& \begin{tabular}[c]{@{}c@{}}   5,394   \\ (21.67   \%)\end{tabular}
& \begin{tabular}[c]{@{}c@{}}   1,153   \\ (30.31   \%)\end{tabular} 
& \begin{tabular}[c]{@{}c@{}}   2,410   \\ (37.78   \%)\end{tabular}
& \begin{tabular}[c]{@{}c@{}}   833     \\ (22.21\%)\end{tabular}    
& \begin{tabular}[c]{@{}c@{}}   35,970     \\ (12.50\%)\end{tabular} 
\\
\hline
\end{tabular}}}
\end{table*}
%
\subsection{Creating Soft-Labels} \label{SEC:Creating_Soft-Labels}
For any image in the training and validation sets of AffectNet, we introduced the \textit{soft-labels} using $SC^{EB}$, the semantic scores of the ensemble of the binary classifiers, and $SC^{AU}$, the semantic scores of AU-based classifier, as follows in Eq.~\ref{eq:soft_labels}:
\begin{align}\label{eq:soft_labels}
\begin{matrix*}[c]
sl(emo_i, img_k) = \frac{1}{2}~ [SC^{EB}_{Mean}(emo_i, img_k) ~ + ~ \\\\ ~~~~~~~~~~~~~~~~~~~~~~~~~~SC^{AU}(emo_i, img_k)], \\\\ SL(img_k) := \{ sl(emo_0, img_k), ..., sl(emo_7, img_k\}.
\end{matrix*}
\end{align}
As Eq.~\ref{eq:soft_labels} expresses, we defined \textit{soft-labels} as a set containing the average of the $SC^{EB}$, and $SC^{AU}$ for each emotion class.

\subsection{Proposed AffectNet+ Dataset} \label{SEC:AffectNet+_Dataset}
The AffectNet+ database is similar to its ancestor, AffectNet, regarding the images in both training and validation sets. Moreover, the original \textit{hard-labels} assigned by human annotators have been retained without modification. By introducing \textit{soft-labels} for each image in the original AffectNet dataset, we propose AffectNet+, which also includes three distinct subsets and supplementary metadata for each image.


\subsubsection{The AffectNet+ Subsets} \label{SEC:AffectNet+_Subsets}
For both the training and validation sets of AffectNet, we introduced 3 different subsets (\textbf{Easy}, \textbf{Challenging}, and \textbf{Difficult}) using the relation between the \textit{soft-labels} and the \textit{hard-labels}.

We defined the Easy subset as the group of images where the emotion with the highest probability in the \textit{soft-label} matches the \textit{hard-label}. Since the highest probability in the \textit{soft-label} aligns with the \textit{hard-label}, it suggests that the facial expression in these images is clear and vivid. As a result, the images in the Easy subset are likely to exhibit distinct and easily recognizable facial expressions.

Next, we introduced the Challenging subset, consisting of the images where the emotion class associated with the \textit{hard-label}, falls within the second or the third-ranked highest probability in the corresponding \textit{soft-label}. To put it simply, although the human-assigned labels (\textit{hard-labels}) may not be the highest probability option for the corresponding \textit{soft-labels}, they still hold a relatively high ranking. Consequently, recognizing the facial expression might be more difficult compared to the images in the Easy subset as the images within this set exhibit complexities or variations that make it less straightforward to identify the primary perceived emotion.

Finally, any images not belonging to either the Easy or the Challenging subsets categorized the Difficult subset. The human-labeled annotations (\textit{hard-labels}) for these images are different from the facial expressions that can be perceived from the corresponding \textit{soft-labels}, indicating the fact that these images represent the most complex and ambiguous cases in terms of recognition of facial expressions. 

Providing the Easy, Challenging, and Difficult subsets allows for the development of different FER models. To illustrate, a classifier trained over the Easy subset can perform more accurately where the facial expressions in the images are high intensity and clearly distinct, while it may face difficulties and confusion in subtle facial images. An ensemble of 3 classifiers, each trained on one of the AffectNet+ subsets, would eventually improve the performance of FER applications specifically in-the-wild settings.

According to Table~\ref{tbl:subset_distribution_affectnet_plus}, which shows a per-class distribution of the AffectNet+ subsets, a significant portion of Happy and Neutral emotions, accounting for 86.25\% and 68.67\%, respectively, are within the Easy set, indicating that the facial expressions associated with these two classes are more obvious compared to other classes. On the contrary, only 12.72\% of Contempt, the least within all the emotion classes, falls under the Easy subset, indicating a high degree of ambiguity associated with this class. 

\begin{figure*}[t]
  \centering
\caption{Distribution of the AffectNet+ sets, including training and validation sets, over different \textit{Easy}, \textit{Challenging}, and \textit{Difficult} subsets.}
\includegraphics[width=1.0\textwidth]{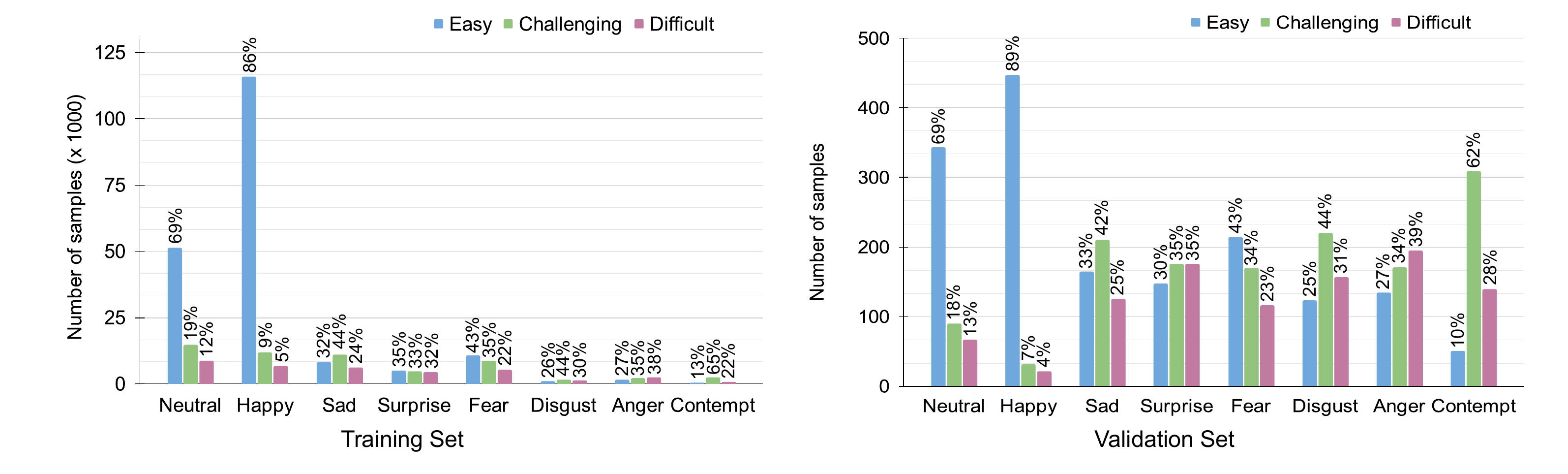}
\label{fig:affectnet+_data_distribution}
\end{figure*}

\subsubsection{Per-Subset Analysis}
Fig.~\ref{fig:affectnet+_data_distribution} depicts the distribution of emotion classes within AffectNet+ subsets. The training set exhibits an imbalance distribution. In the Easy subset, there exists the maximum number of Happy and Neutral facial images, while the Contempt and Disgust expressions are the least represented images. Likewise, this trend exists in the Challenging and Difficult sets. The imbalanced training set causes challenges for training Hard-FER models, addressed using a combination of up-sampling and weighted loss. However, Soft-FER models require no such adjustments during training. In addition, the class imbalance is apparent in the validation set, highlighting the need for reporting metrics like F-1 score and average accuracy, alongside accuracy.

\subsubsection{Per-Expression Analysis}
Fig.~\ref{fig:affectnet+_data_distribution} also shows the distribution of images in the AffectNet+ training and validation sets across emotion classes and subsets. In both the training and validation sets, the Happy and Neutral classes were dominant in the Easy subset. Conversely, the Contempt and Disgust classes have minimal representation in the Easy set but show higher proportions in the Challenging set, indicating that ambiguity exists in the perception of this emotion class for humans. Overall, for most emotions, over two-thirds of the images fall within the Challenging and Difficult sets, highlighting complexities in interpreting facial expressions. Consequently, while the traditional hard-label struggles to represent the full expression spectrum, the proposed \textit{soft-labels} provide information regarding the combination of various expressions with different intensities.

\subsubsection{AffectNet+ Metadata}
To further enrich the AffectNet+ dataset, for each image in the training and validation set, we provided \textit{gender}, \textit{age}, \textit{ethnicity}, two new sets of \textit{facial landmark points} (68-point and 28-point), and \textit{head pose} as metadata, using pre-trained deep learning-based models. For \textit{gender} classification, we used the model proposed by Rothe~\etal~\cite{Rothe-IJCV-2018}. For both \textit{age}, and \textit{ethnicity} classification, we utilized the model introduced by Serengil~\etal~\cite{serengil2021lightface}. For both 68-point and 28-point landmark localization, we used the model provided by Fard and Mahoor~\cite{fard2022acr}. We utilized ASMNet by Fard~\etal~\cite{fard2021asmnet} for estimating head pose as a combination of \textit{yaw}, \textit{pitch}, and \textit{roll}. For age, the corresponding age detector~\cite{serengil2021lightface} predicts a numerical value. As for \textit{gender} classification, the classifier~\cite{Rothe-IJCV-2018} assigns the class labels Man and Woman to each image. Likewise, for \textit{ethnicity}, the classifier~\cite{Rothe-IJCV-2018} assigns Indian, Black, White, Middle-Eastern, and Hispanic to each image. Refer to Supplementary Materials for more details. 


\section{Experimental Results} \label{SEC:EXPERIMENTAL_RESULTS}
In this section, we first elaborate on the ensemble of binary classifiers, explain the implementation detail and evaluation method, and analyze the models' performance. Then, we assess the performance of our proposed AU-based classifier and review the details of its implementation. Finally, we introduce new baseline models for both Hard-FER and Soft-FER on each subset of the AffectNet+ dataset.

\begin{table}[b] 
\caption{Accuracy and the average accuracy, over each expression, using ensemble of binary classifiers
(EBC), over test-MAS (in \%). The average accuracy ($\overline{Acc}$) shows the average of the true positives and true negatives, based on Eq.~\ref{eq:SC_bin_cls}.}
\label{tbl:bin_eval_test_MAS}
\centering
\small
\resizebox{0.5\textwidth}{!}
{{
\begin{tabular}{lcccccccc}
\hline
& Neutral & Happy & Sad  & Surprise & Fear & Disgust & Anger & Contempt \\ \hline
\multicolumn{9}{c}{\textbf{ResNet-50\cite{he2016deep}}} \\ \hdashline
Acc     & 81.42    & 84.14  & 80.79 & 86.90     & 87.54 & 88.91    & 81.48  & 77.47     \\
$\overline{Acc}$ & 79.48    & 87.13  & 77.00 & 85.64     & 83.85 & 80.37    & 83.92  & 65.23     \\ \hline
\multicolumn{9}{c}{\textbf{EfficientNet-B3\cite{tan2019efficientnet}}} \\ \hdashline
Acc     & 82.92    & 84.46  & 81.29 & 88.02     & 86.85 & 87.08    & 84.64  & 75.41     \\
$\overline{Acc}$ & 82.91    & 87.33  & 79.88 & 86.34     & 82.56 & 85.78    & 82.59  & 69.60     \\ \hline
\multicolumn{9}{c}{\textbf{XceptionNet\cite{chollet2017xception}}} \\ \hdashline
Acc     & 81.31    & 82.79  & 82.56 & 89.43     & 88.38 & 91.47    & 88.01  & 84.64     \\
$\overline{Acc}$ & 81.55    & 88.04  & 82.76 & 87.48     & 84.79 & 87.41    & 87.63  & 65.51     \\ \hline
\multicolumn{9}{c}{\textbf{Ensemble of Binary Classifiers}} \\ \hdashline
Acc     & 84.38    & 84.09  & 88.79  & 89.63     & 92.77 & 91.62    & 84.57  & 87.93     \\
$\overline{Acc}$ & 88.52    & 87.88  & 84.64  & 91.10     & 88.62 & 86.56    & 85.19  & 78.51     \\ \hline
\end{tabular}
}}
\end{table}

\begin{table}[b] 
\caption{Accuracy and average accuracy over each expression, using AU-base classifier, over test-MAS (in \%). The average accuracy ($\overline{Acc}$) is calculated based on Eq.~\ref{eq:SC_bin_cls}.}
\label{tbl:action_unit_eval_test_MAS}
\centering
\small
\resizebox{0.50\textwidth}{!}
{{
\begin{tabular}{lcccccccc}
\hline
 & Neutral & Happy & Sad   & Surprise & Fear & Disgust & Anger & Contempt \\ \hline
Acc     & 84.93    & 82.03  & 67.00    & 75.89     & 31.81 & 80.92    & 50.48  & 70.33     \\
$\overline{Acc}$ & 88.43    & 87.10  & 75.57  & 85.38     & 60.14 & 81.43    & 67.41  & 77.38     \\ \hline
\end{tabular}
}}
\end{table}

\subsection{Ensemble of Binary Classifiers Results} \label{SEC:EXPERIMENTAL_RESULTS_ens_cls}
~~~
\textbf{Training:} We selected ResNet-50~\cite{he2016deep}, EfficientNet-B3~\cite{tan2019efficientnet}, and XceptionNet~\cite{chollet2017xception} as our backbone models. We trained each model for every emotion class individually (one-vs-rest) using the train-MAS subset. With these three backbone models and eight expressions, we generated a total of 24 different decision-makers. For the training step, we followed the methodology described in Sec.~\ref{SEC:METHODOLOGY_ens_cls_train}. To this end, we split data into the positive and negative samples. Positives were the samples with a specific label (like Happy), and negatives were the rest. To train each model, our method re-scaled each image to the size of $224 \times 224$ and utilized the Adam optimizer~\cite{kingma2014adam} with $learning-rate = 10^{-3}$, $\beta_1 = 0.9$, $\beta_2 = 0.999$, and $decay = 10^{-5}$, for 25 epochs with a batch size of 50. We implemented our models using TensorFlow and ran them on Nvidia GPUs.

\textbf{Test:} To evaluate the performance of our trained binary classifiers, we leveraged the test-MAS subset. This set included 800 uniform samples, therefore for every binary classifier (like Happy), we had 100 positive and 700 negative samples. As mentioned earlier, we ensembled three models, including ResNet-50~\cite{he2016deep}, EfficientNet-B3~\cite{tan2019efficientnet}, and XceptionNet~\cite{chollet2017xception} models. Different popular metrics, including precision, recall, F-1 score, accuracy, as well as average accuracy (See Eq.~\ref{eq:SC_bin_cls}), were used for evaluating the ensemble of binary classifier (EBC) models. A summary of these metrics is shown in Table~\ref{tbl:bin_eval_test_MAS}, while the full table is provided in Supplementary Materials. 

The reported accuracy (shown by $Acc$) in Table~\ref{tbl:bin_eval_test_MAS} depicts that in our one-vs-rest model training, we could reach high accuracies for all the expressions, which is an indication of our models' robustness. All three classifiers significantly boosted the accuracy of the least provided samples (like Contempt). This table highlights that the $Acc$ varied between 75\% and 92\% for all the classifier models, per expression. Meanwhile, the standard deviation of the classifiers' $Acc$ for all the expressions was 3.95\%, 4.07\%, and 3.72\%, for ResNet-50~\cite{he2016deep}, EfficientNet-B3~\cite{tan2019efficientnet}, and XceptionNet~\cite{chollet2017xception}, respectively. The $Acc$ in the last section of this table demonstrates the results of an ensemble of three aforementioned models, where the $Acc$ per expression changed in the higher range of 87\% to 93\%, and the standard deviation was lower than each of the three models (3.36\%). 

On the other hand, the average accuracy (shown by $\overline{Acc}$) tried to highlight the impact of the imbalance distribution of the validation set (100 positive samples versus 700 negative samples). The difference between $Acc$ and $\overline{Acc}$ of the Neutral, Happy, Sad, Surprise, Fear, and Anger expressions was not eye-catching. This fact, bolds the low impact of the imbalance data distribution on our training method. Notably, the highest impact of the imbalance distribution was shown on Disgust and Contempt expressions on the three ResNet-50~\cite{he2016deep}, EfficientNet-B3~\cite{tan2019efficientnet}, and XceptionNet~\cite{chollet2017xception} models. However, even the reported $\overline{Acc}$ on these two expressions was considerable for all of the models. Thanks to the method we utilized for the ensemble of a binary classifier, we boosted $\overline{Acc}$ of Disgust and Contempt expressions to 86.56\% and 78.51\%, respectively. This fact demonstrates the effect of our ensemble model on the imbalanced data. In summary, analyzing Table~\ref{tbl:bin_eval_test_MAS} illustrates the reliability of the proposed ensemble of binary classifiers (EBC) model, with high accuracies and low standard deviations for all the expressions, useful for the expression classification and \textit{soft-labeling}.

\subsection{Action Unit (AU)-Based Classifier Results} 
\label{SEC:EXPERIMENTAL_RESULTS_au_cls}
~
\textbf{Training:} We selected ResNet-50~\cite{he2016deep} as the backbone of our AU-based classifier. As described in Sec.~\ref{SEC:METHODOLOGY_au_cls}, we modified the last layer of ResNet-50~\cite{he2016deep}, such that the model has two outputs, the binary probabilities, and the AU-based representation vector. For each emotion class, we trained the corresponding model individually, using the train-MAS subset of images. The binary probability refers to the probability distribution over different classes, while the AU-based representation indicates the intensity of an AU in an image. We trained the models using the images with a size of $224 \times 224$ pixels. We used the Adam optimizer~\cite{kingma2014adam} with $learning$-$rate = 10^{-4}$, $\beta_1 = 0.9$, $\beta_2 = 0.999$, and $decay = 10^{-6}$, for 40 epochs with a batch size of 50. We implemented these models in Tensorflow with the same GPU used for binary classifiers.

\textbf{Test:} As described in Sec.~\ref{SEC:METHODOLOGY_au_cls_conf_score}, we first utilized a post-processing algorithm to convert the AU-based representation vector to a binary probability vector. Next, we took the average of the probability vectors of the binary classification task and the AU-based representation vector task, as the final decision of each model. 

Similar to the ensemble of binary classifiers, we evaluated our AU-based classifier over test-MAS. Precision, recall, F-1 score, accuracy, and average accuracy were the selected metrics for this analysis. Table~\ref{tbl:action_unit_eval_test_MAS} shows the accuracy and average accuracy of the AU-based classifier. To see the full results, refer to Supplementary Materials. AU-based classifier worked well for the expressions Neutral, Happy, Sad, Surprise, Disgust, and Contempt. The accuracy (shown by $Acc$) over these expressions was in the range of 67\% to 85\%. However, the accuracy for two expressions, Fear and Anger, was lower than the other expressions. The high number of common action units between the Fear expression and other expressions was the reason for its lowest accuracy among all the expressions. There were 4 common action units between the two expressions Fear and Anger, which were highly activated in both the Fear and Anger facial samples. Fear also had 5 common action units with Surprise, which were less activated in the facial samples, and affected the accuracy of the Fear expression. On the other hand, the average accuracy (shown by $\overline{Acc}$) of the AU-based classifiers was higher than 60\% for all the expressions. This table demonstrates that with a subtle analysis of the facial expressions, using their action units, we could extract valuable information for the expression classification and \textit{soft-labeling}.

To evaluate the role of the AU-based classifier in the final decision-making, we conducted an experiment. We trained a model (ResNet-50~\cite{he2016deep}) to label images without and with an AU-based classifier. This experiment showed that for all the expressions the accuracy increased when we added an AU-based classifier to our baseline model. The progress over the average accuracy was eye-catching (up to 10\%). To see the table of this experiment refer to Supplementary Materials. 

\subsection{Baseline Models for AffectNet+} \label{SEC:EXPERIMENTAL_RESULTS_baseline}
In this section, we provide a set of new baselines for both Hard-FER and Soft-FER methods on AffectNet+. We used ResNet-50~\cite{he2016deep} as the backbone for both methods. We trained the baseline models on the training set of AffectNet+, and assessed their accuracy and performance on the validation set of AffectNet+. 
\begin{table}[b] 
\caption{Accuracy and average accuracy of Hard-FER on baseline model (ResNet-50\cite{he2016deep}) over AffectNet+.}
\label{tbl:resnet_hard_label_accuracy}
\centering
\small
\resizebox{0.40\textwidth}{!}
{{
\begin{tabular}{lcccc}
\hline
        & All   & Easy & Challenging & Difficult  \\ \hline 
Acc (\%)    & 52.06 & 85.86  & 51.62       & 34.34 \\
$\overline{Acc}$ (\%) & 52.04 & 78.13  & 52.38       & 39.15 \\ \hline 
\end{tabular}}}
\end{table}

\textbf{Baselines for Hard-FER:}\label{SEC:EXPERIMENTAL_RESULTS_baseline_HFER}
For Hard-FER, we trained our baseline model using the \textit{hard-labels}, provided by the human annotators. For each subset of AffectNet+, we trained one baseline model and evaluated its accuracy and performance on its corresponding subset in the validation set. Table~\ref{tbl:resnet_hard_label_accuracy} shows the accuracy and the average accuracy of Hard-FER baseline models. According to Table~\ref{tbl:resnet_hard_label_accuracy}, the baseline model achieved the highest accuracy (85.86\%) on the Easy subset, by far greater than the accuracy on the Challenging and Difficult subsets, 51.62\% and 34.34\% respectively. These results are expected since as elaborated in Sec.~\ref{SEC:AffectNet+_Subsets}, the subsets within AffectNet+ vary in terms of facial expression intensity and ambiguity, which directly influences the accuracy of FER. The images within the Easy subset tend to have high-intensity facial expressions, while the faces in the Challenging and Difficult sets tend to have less intense, more ambiguous expressions.

Table~\ref{tbl:resnet_hard_label_prec_rec_f1} shows precision, recall, and F-1 score for each subset of AffectNet+. This table reveals that, over all the sample data, the baseline model achieved the highest F-1 score for the Happy class (66.05\%), and the lowest for Contempt (25.32\%). We witnessed a similar pattern for the Easy subset, where the F-1 scores for Happy and Contempt classes are 95.55\% and 51.94\%, respectively. However, for the Challenging and Difficult subsets, the lowest F-1 score was achieved for the Neutral and Happy classes. It can be concluded that although Happy and Neutral were among the most obvious and less ambiguous emotions for FER, in subtle cases can still be extremely difficult to recognize these emotions. Overall, as we expected, the F-1 score reduced on the Challenging and Difficult subsets, in comparison with the Easy subset.

\begin{figure*}[t]
  \centering
  \includegraphics[width=1.0\textwidth]{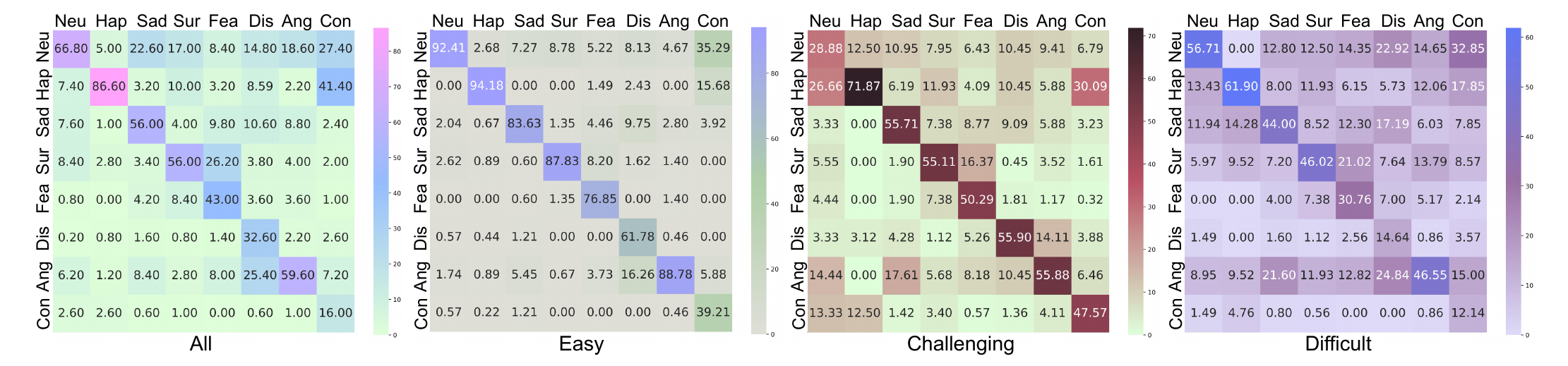}
  \caption{Confusion matrix of the baseline model (ResNet-50~\cite{he2016deep}) for every subset of AffectNet+ (\textit{Easy}, \textit{Challenging}, and \textit{Difficult}). The baseline model is trained over any subset, separately. Then, the models are evaluated over all the samples in the evaluation set, regardless of their subset.}
  \label{fig:resnet_conf}
\end{figure*}
%

Fig.~\ref{fig:resnet_conf} shows the confusion matrices of the baseline model for every subset of AffectNet+. The model faced the least confusion on the Easy set, while the highest level of confusion occurred on the Difficult set. This figure indicates that images within the former set include obvious facial expressions that are unambiguous and easy to recognize, whereas the latter comprises images with less distinct facial expressions. Furthermore, considering all the subsets, the highest degree of confusion happened between the facial expression of Contempt and either Neutral or Happy, and the second highest level of confusion was between Disgust and Anger. For the Challenging and Difficult sets, we observed a high level of confusion between the facial expressions of Sad and Anger, as well as between Surprise and Fear. The high level of confusion between facial expressions associated with specific emotions, indicating the intra-class variations, as well as inter-class similarities, clearly explains how Hard-FER results in an inaccurate FER model and illustrates the effectiveness of our proposed Soft-FER method.

\textbf{Baselines for Soft-FER}\label{SEC:EXPERIMENTAL_RESULTS_baseline_SFER}
As described in Sec.~\ref{SEC:METHODOLOGY_SFER}, in our proposed Soft-FER methodology, the neural network trained to predict the probability scores for facial expressions associated with each individual emotion class. Since the prediction of \textit{soft-labels} is a regression task, we utilized mean error, failure rate, and Area Under the Cumulative Errors Distribution curve~\cite{yang2015empirical} as the evaluation metrics. 

\begin{table}[b] 
\caption{Per-class precision, recall, and F-1 score of Hard-FER baseline model, ResNet-50\cite{he2016deep}, for each expression on the AffectNet+ dataset (in \%).}
\label{tbl:resnet_hard_label_prec_rec_f1}
\centering
\small

\resizebox{0.50\textwidth}{!}
{{
\begin{tabular}{lcccccccc}
\hline
     & Neutral & Happy & Sad   & Surprise & Anger & Disgust & Fear  & Contempt \\ \hline
     & \multicolumn{8}{c}{\textbf{All}}                                                 \\ \hdashline
Prec & 36.97   & 53.33 & 55.82 & 52.45    & 66.56 & 77.25   & 50.16 & 65.00    \\
Rec  & 66.66   & 86.74 & 55.82 & 56.04    & 43.00 & 32.79   & 59.55 & 15.72    \\
F-1  & 47.56   & 66.05 & 55.82 & 54.19    & 52.24 & 46.04   & 54.46 & 25.32    \\ \hline
     & \multicolumn{8}{c}{\textbf{Easy}}                                              \\ \hdashline
Prec & 79.39   & 96.99 & 78.40 & 81.13    & 94.49 & 91.56   & 80.16 & 76.92    \\
Rec  & 92.39   & 94.15 & 84.14 & 87.75    & 76.86 & 61.78   & 88.78 & 39.21    \\
F-1  & 85.40   & 95.55 & 81.17 & 84.31    & 84.77 & 73.78   & 84.25 & 51.94    \\ \hline
     & \multicolumn{8}{c}{\textbf{Challenging}}                                         \\ \hdashline
Prec & 17.64   & 10.95 & 62.36 & 66.66    & 75.22 & 67.22   & 45.23 & 80.00    \\
Rec  & 27.90   & 71.87 & 56.31 & 54.85    & 49.10 & 56.01   & 56.21 & 46.82    \\
F-1  & 21.62   & 19.00 & 59.18 & 60.18    & 59.42 & 61.11   & 50.13 & 59.07    \\ \hline
     & \multicolumn{8}{c}{\textbf{Difficult}}                                                \\ \hdashline
Prec & 18.81   & 11.50 & 37.16 & 45.76    & 61.45 & 58.97   & 27.97 & 77.27    \\
Rec  & 56.71   & 61.90 & 44.00 & 46.28    & 30.72 & 14.83   & 46.55 & 12.23    \\
F-1  & 28.25   & 19.40 & 40.29 & 46.02    & 40.97 & 23.71   & 34.95 & 21.11    \\ \hline
\end{tabular}
}}
\end{table}

To better evaluate the model performance, we proposed a weighted error mechanism to measure the error between the ground truth and the generated \textit{soft-labels}. We assigned a weight to each element of an arbitrary ground truth \textit{soft-label}, based on its relative magnitudes. To clarify, the weight associated with the $i^{th}$ element is proportional to its relative magnitudes, such that the largest element will be receiving a weight of 1, the second largest element, a weight of $\frac{1}{2}$, and so on (the weight $\frac{1}{8}$ will be assigned to the smallest element). The weighting mechanism ensured that the elements with higher values in a ground truth \textit{soft-label} are considered more important compared to the elements with lower values. We calculated the Weighted Mean Absolute Error (W-MAE) as follows in Eq.~\ref{eq:soft_labels}:
\begin{align}\label{eq:mean_error}
\begin{matrix*}[r]
\text{W-MAE} = \frac{100}{N \times n} \sum_{k=0}^{N} \sum_{i=0}^{n} w^{i}_k |SL^{i}_k - \hat{SL^{i}_k}|,
\end{matrix*}
\end{align}
\begin{table}[b!] 
\caption{Weighted failure-rate (W-FR) and weighted mean average error (W-MAE) of Soft-FER baseline model (ResNet-50\cite{he2016deep}) on AffectNet+.}
\label{tbl:resnet_soft_label_error}
\centering
\small
\resizebox{0.40\textwidth}{!}
{{
\begin{tabular}{lcccc}
\hline
                  & All   & Easy & Challenging & Difficult  \\ \hline
W-FR (\%)    & 10.85     & 8.00     & 11.90       & 18.66 \\
W-MAE (\%)  & 17.30      & 15.43      & 18.54        & 21.21  \\
\hline
\end{tabular}}}
\end{table}

where $N$ is the number of images in the validation set, $n$ is the number of emotions in the $EMOTIONS$ set, $SL^{i}_k$ and $\hat{SL^{i}_k}$ are the $i^{th}$ elements of the ground truth, and the predicted \textit{soft-labels}, respectively, associated with the $k^{th}$ image. Finally, $w^{i}_k$ is the weight of the $i^{th}$ element of the \textit{soft-label} corresponding to the $k^{th}$ image.

Building upon W-MAE, we proposed the Weighted Failure Rate (W-FR), a metric to show the robustness of the models. To calculate the W-FR, first, we defined a threshold, called $\epsilon$. Then, an individual prediction was considered a \textit{failure} if the weighted error between the ground truth and its corresponding predicted \textit{soft-label} was greater than $\epsilon=0.3$. W-FR is defined as the portion of these failures among all predictions.


Table~\ref{tbl:resnet_soft_label_error} shows the W-MAE and W-FR for each subset of AffectNet+. Similar to Hard-FER, W-MAE, and W-FR are small for the Easy subset (17.30\% and 10.58\%, respectively), and large for the Difficult subset (21.21\% and 18.66\%, respectively), representing the degree of difficulty of facial expression recognition for each subset. 

\begin{table}[b!] 
\caption{Per-class weighted failure-rate (W-FR), and weighted mean average error (W-MAE) of Soft-FER baseline model (ResNet-50\cite{he2016deep}) on AffectNet+ (in \%).}
\label{tbl:resnet_soft_label_error_on_expression}
\centering
\small
\resizebox{0.45\textwidth}{!}
{{
\begin{tabular}{llcccc}
\hline
                          &     & All   & Easy & Challenging & Difficult  \\  \hline
\multirow{2}{*}{Neutral}  & W-FR  & 9.60  & 5.54   & 21.11       & 17.91 \\
                          & W-MAE & 17.46 & 15.29  & 22.60       & 23.09 \\
                          \hdashline
\multirow{2}{*}{Happy}    & W-FR  & 3.60  & 2.24   & 12.50       & 19.05 \\
                          & w-MAE & 12.59 & 11.27  & 18.24       & 22.84 \\
                          \hdashline
\multirow{2}{*}{Sad}      & W-FR  & 11.60 & 8.48   & 10.48       & 19.20 \\
                          & W-MAE & 18.14 & 16.85  & 17.07       & 20.39 \\
                          \hdashline
\multirow{2}{*}{Surprise} & W-FR  & 12.60 & 13.51  & 13.64       & 18.18 \\
                          & W-MAE & 18.25 & 17.60  & 18.43       & 20.43 \\
                          \hdashline
\multirow{2}{*}{Fear}     & W-FR  & 17.80 & 20.14  & 11.70       & 25.13 \\
                          & W-MAE & 19.42 & 18.71  & 18.92       & 22.43 \\
                          \hdashline
\multirow{2}{*}{Disgust}  & W-FR  & 12.00 & 16.26  & 15.00       & 19.75 \\
                          & W-MAE & 18.14 & 18.18  & 19.55       & 21.86 \\
                          \hdashline
\multirow{2}{*}{Anger}    & W-FR  & 11.20 & 8.41   & 12.35       & 12.07 \\
                          & W-MAE & 17.80 & 17.12  & 19.29       & 19.39 \\
                          \hdashline
\multirow{2}{*}{Contempt} & W-FR  & 8.40  & 3.92   & 6.80        & 14.29 \\
                          & W-MAE & 16.55 & 19.56  & 17.12       & 20.84 \\
                          \hline
\end{tabular}
}}
\end{table}

In Table~\ref{tbl:resnet_soft_label_error_on_expression}, we provided a per-emotion analysis of the performance of the baseline model in Soft-FER. Overall, for the Easy set, we observed the lowest W-FR and W-MAE values. The highest values were observed in the Difficult set. Considering all the samples in the validation set (marked as \textit{All} in Table~\ref{tbl:resnet_soft_label_error_on_expression}), the baseline model performed the best in terms of recognizing Happy expression, and the worst in terms of recognizing Fear and Disgust expressions. For the Easy set, the baseline model achieved the lowest W-FR and W-MAE on Happy, Neutral, and Contempt. Contrary to Hard-FER, where the baseline model has a high confusion rate between the Neutral and the Contempt expressions, Soft-FER showed an improved performance. This occurred because Soft-FER considered each emotion class individually and predicted the probability scores associated with each class given a facial image.   

\section{Subjective Evaluation of Soft-Lables} \label{SEC:SUBJECTIVE_EVALUATION}
The concept of \textit{soft-labeling} offers a more nuanced representation of data and helps soften the classification boundaries in models. It could also provide insights into compound labeling, as noted by many recent studies. In addition to model-based evaluations, human assessment is crucial for evaluating the potential benefits of \textit{soft-labeling}. We thus conducted an experiment with human participants to compare the utility of soft and hard labels for accurately reflecting people's subjective perception of emotion on others' faces. 

\begin{figure*}[t]
  \centering
\caption{The results of the subjective tests highlight the importance of the \textit{soft-labeling} approach from a human perspective. Subfigure (a) demonstrates that subjects preferred \textit{soft-labels} over \textit{hard-labels} to describe the images. Subfigure (b) shows that subjects were able to distinguish between accurate and random \textit{soft-labels} for individual images. Subfigure (c) depicts the self-agreement of each participant and the inter-agreement between them. Self-agreement is indicated by the nodes, while the edges represent pair-wise inter-agreement.}
\includegraphics[width=0.9\textwidth]{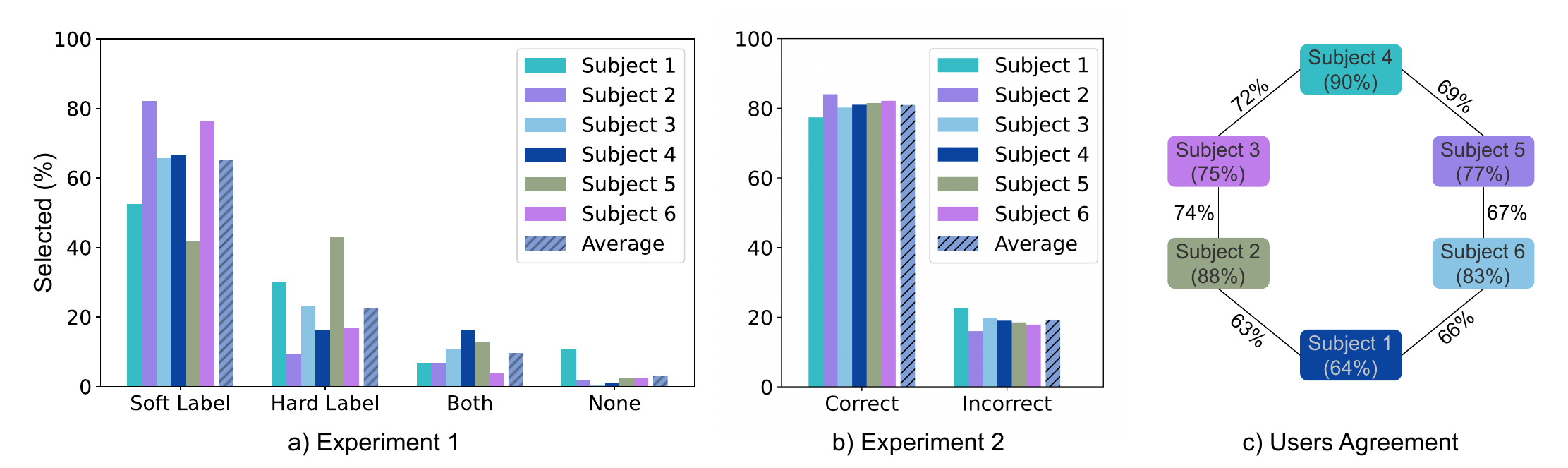}
\label{fig:subjective_test}.
\end{figure*}

\subsection{Subjective Test Design}
\label{SEC:SUBJECTIVE_TEST_DESIGN}
There are two key questions about the \textit{soft-labeling} approach compared to the traditional \textit{hard-labeling} approach: from a human perspective, \textbf{1)} which approach is more informative for explaining the expressions of a facial image, and \textbf{2)} how accurately can \textit{soft-label} describe the expressions of a facial image.

We selected 6 students from a diverse pool of candidates, ensuring a range of ages, genders, and racial backgrounds. Their task was to review a large set of facial images from the AffectNet+ evaluation set and respond to two key questions. The evaluation set consisted of 500 images for each of the eight expression categories, totaling 4000 images. Since there were 2 questions per image, this resulted in 8000 questions. Additionally, 30\% of these questions were repeated for reliability: 20\% involved self-evaluation, and 10\% were for circular user agreement, where each user was compared with 2 other users. In total, we had 10,406 questions, randomly and equally distributed among the experimenters.

For the first experiment, we showed each experimenter a random facial image and asked him/her to select the best facial image descriptor among \textit{hard-labels}, \textit{soft-labels}, both, and none. In the other experiment, we showed each experimenter a facial image accompanied by two \textit{soft-labels} and asked him/her to select the \textit{soft-label} related to the facial image, among the related and a randomly selected \textit{soft-label}. The \textit{soft-labels} were created using the approach described in Section~\ref{SEC:Creating_Soft-Labels}, while the \textit{hard-labels} were the original human annotations from AffectNet. To find more details on these two experiments refer to Supplementary Materials.

All participants were students from the University of Denver, aged 20-45 years. The study was conducted under an approved IRB, and the students provided consent. The group included 4 males and 2 females, representing diverse racial backgrounds: Asians (2 students), Hispanic-Latino, Caucasian, Middle-Eastern, and White-Asian. We organized a training session for all participants to review universal facial expressions and their associated facial indicators and appearances. To ensure they were adequately trained, we asked each participant to label 40 images from our dataset and evaluated their performance against the image labels. A 75\% agreement with the labels was required to qualify, and all participants passed this exam before beginning the main experiment. Each participant was then randomly assigned approximately 1,735 images and given 7 days to complete the task.

\subsection{Subjective Test Analysis} \label{SEC:SUBJECTIVE_TEST_ANALYSIS}
Fig.~\ref{fig:subjective_test} shows the results of two experiments, and the percentage of agreement between subjects. As Fig.~\ref{fig:subjective_test}-a demonstrates, on average, human subjects preferred \textit{soft-labels} in 65\% of the first experiment, compared to only 22\% for \textit{hard-labels}. Additionally, 5 out of 6 participants selected \textit{soft-labels} as the best descriptor overall. The results indicate that, on average, humans preferred \textit{soft-labels} over traditional \textit{hard-labels} as the better image descriptor.

Fig.~\ref{fig:subjective_test}-b evaluates the reliability of \textit{soft-labels}, where in some test cases only the intensity of the expression posed a challenge for the experimenters. The results illustrate tht, given an accurate versus a random \textit{soft-label}, participants could identify the accurate \textit{soft-label} in 81\% of the questions. The accuracy of each participant in the second experiment varied between 77\% and 84\%. 

Finally, Fig.~\ref{fig:subjective_test}-c examines how accurately the participants answered the questions. We asked 30\% of the images more than once to evaluate the self-agreement and pairwise agreement between the participants. The average accuracy across all the participants was 80\%, with a maximum of 90\% and a minimum of 64\%. These results show that the users' agreement was also notable, as participants largely agreed on the common questions, with an average agreement of 69\%, which is high for FER tasks.

The results of these experiments confirmed that, from a human perspective, the concept of \textit{soft-labeling} provides a more accurate and intuitive description of facial expressions. This is particularly relevant as many facial images convey more than one distinct expression with varying intensities.

\section{Future Research Direction} \label{SEC:FUTURE_RESEARCH}
Mixed facial expressions are common in real-life emotional displays, making them important to consider when studying both human and computer-based recognition of facial affect. Many recent studies in FER thus focus on compound-labeling and \textit{soft-labeling}. Applying \textit{soft-labels} allows for more nuanced FER, more flexibly responding to the true complexity of facial expressions as they are produced in the wild, with subtlety and sometimes multiple emotions conveyed at the same time. AffectNet+ can open some windows to the problems that need compound-labeling or \textit{soft-labeling}. The complexity of FER datasets can be originated from extrinsic and intrinsic challenges. Extrinsic challenges originate from extrinsic factors such as illumination, camera quality, and query type (for in-the-wild datasets). Besides, intrinsic challenges occur because of noisy labels, relative relation of expressions, intensity of expressions, diversity of the facial samples, head pose, eye movements, etc. AffectNet+ provides the opportunity to focus on some of these issues. In continuing, we introduce some of the future research directions using AffectNet+.
\begin{itemize}
\item AffectNet+ is a source for research on quantifying uncertainty in \textit{soft-label} prediction. 
\item Multi-labeling is another feature proposed by AffectNet+, where it allows FER models to predict even more than two expressions from a facial image.
\item \textit{Soft-labels} provided in AffectNet+ can help future studies to reduce the effect of noisy labels.
\item Using AffectNet+ we can find smoother decision boundaries. Therefore, studying generalization over \textit{soft-labels} and comparing them with \textit{hard-labels} could be another future research direction using this dataset. 
\item AffectNet+ can be a source to study imbalanced data and provide solutions to this challenge. This dataset can also be considered a multi-expression dataset, where the data distribution is less imbalanced.
\item AffectNet+ could be a source for domain adaptation in FER. Domain adaptation is an open problem in machine learning. Transferring knowledge from models trained on AffectNet+ to video-based dynamic facial tracking tasks is another potential research topic.
\item Interpretability studies of FER models are possible using AffectNet+. Joint \textit{soft-label} and \textit{hard-label} model training can maintain the interpretability of one-hot training while utilizing smoother expression margins at the same time.  
\item AffectNet+ provides the intensity of \textit{soft-label} expressions; therefore, designing FER models and loss functions that consider the labels and their intensity during training is effective in FER studies. 
\item The three subsets of AffectNet+ provide the opportunity to train and combine different models for each subset with various loss functions and regularizers.
\item Metadata provided in AffectNet+ is a valuable source for coping with imbalanced data in FER. Additionally, this dataset is practical for data augmentation and self-supervised learning.
\end{itemize}

The aforementioned research problems highlight the importance and essence of AffectNet+ over facial expression recognition tasks. Best of our knowledge, AffectNet+ is the largest human-annotated in-the-wild dataset, accompanied by \textit{soft-labels} and metadata, for the next studies on facial expression recognition.  
\section{Conclusion} \label{SEC:CONCLUSION}
Automated Facial Expression Recognition plays a crucial role in understanding human emotions and has diverse applications in healthcare, autonomous driving, and education. The advent of deep learning techniques, such as Convolutional Neural Networks and Vision-Transformers, has significantly improved the accuracy of FER methods. However, FER remains challenging due to intra-class variations, inter-class similarities, and cultural differences in perceiving and judging facial expressions. The existing FER datasets suffer from limited annotations, noisy labels, and biased models, hindering the development of robust and reliable FER systems.

To alleviate these challenges, we proposed Soft-FER, a novel approach for FER, alongside the traditional (Hard-FER). We introduced the concept of \textit{soft-labels} in the FER datasets, which provides the probability score of facial expression existence for each emotion class in an arbitrary image. Compared to the traditional \textit{hard-labels}, where we assign only one label to images, \textit{soft-labels} are more explanatory, enabling a more comprehensive and nuanced representation of emotions and resulting in the development of more accurate aromatic FER solutions. We proposed two novel methods, an ensemble of binary classifiers and an AU-based classifier, for an accurate calculation of \textit{soft-labels} for each image in AffectNet. 

Building upon AffectNet, we proposed the AffectNet+ dataset, by adding \textit{soft-labels} to each image and providing additional metadata. Moreover, we introduced 3 new subsets (i.e., Easy, Challenging, and Difficult subsets) to AffectNet+, based on the difficulty of the recognition of facial expressions. AffectNet+ has the potential to be utilized to enhance the performance and robustness of FER systems, resulting in a better interpretation of human facial expressions.

\bibliographystyle{IEEEtran}
\bibliography{main.bib}
\vspace{-2cm}
\begin{IEEEbiography}[{\includegraphics[width=1in,height=1.25in,clip,keepaspectratio]{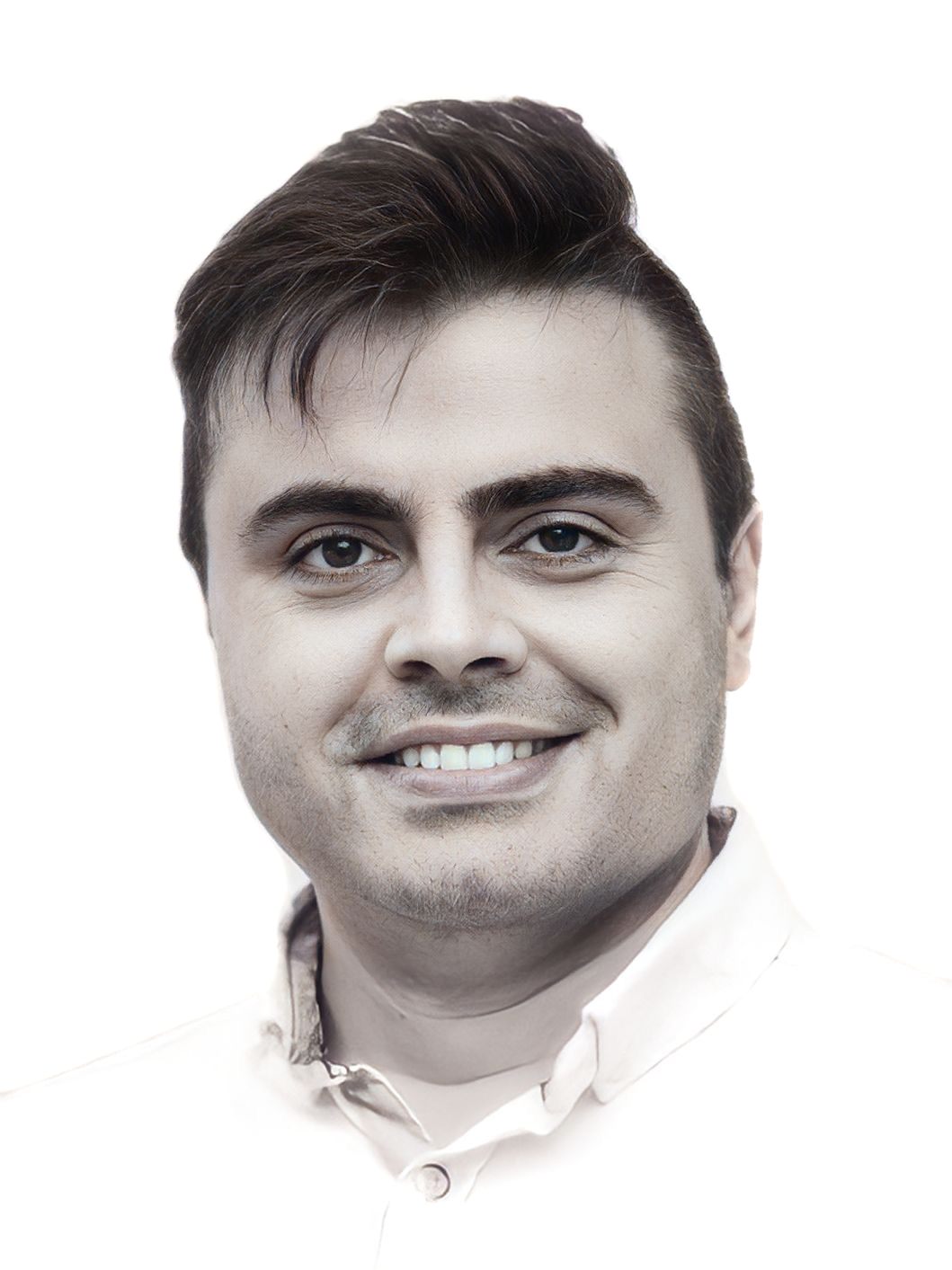}}]{Ali Pourramezan Fard} received an MS degree in Computer Engineering from Iran University of Science and Technology, Tehran, Iran, in 2015. He is currently pursuing his Ph.D. degree in Electrical \& Computer engineering at the University of Denver. His research interests include computer vision, machine learning, and deep neural networks, especially in face alignment, and facial expression analysis.
\end{IEEEbiography}

\vspace{-2.2cm}
\begin{IEEEbiography}[{\includegraphics[width=1in,height=1.25in,clip,keepaspectratio]{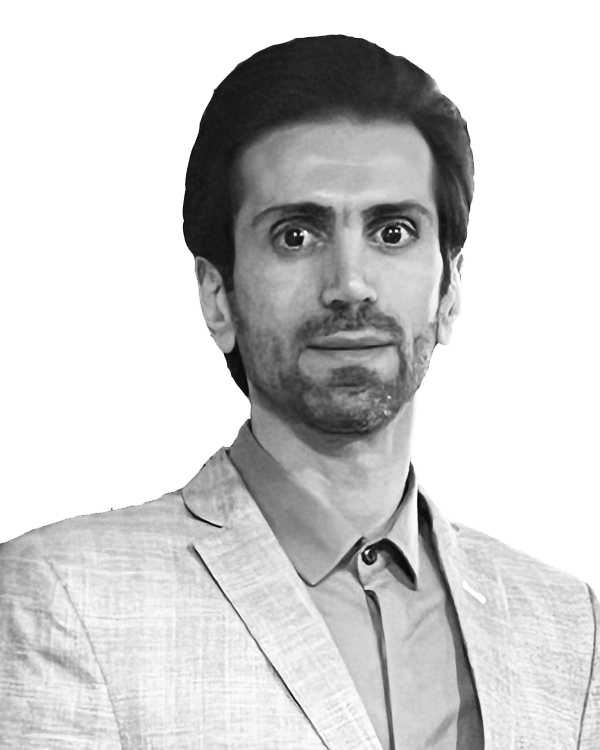}}]{Mohammad~Mehdi~Hosseini} received an MS degree in Computer Engineering from Sharif University of Technology, Iran, in 2015. He is currently pursuing his Ph.D. in Electrical \& Computer Engineering at the University of Denver.  His research interests include pattern recognition, machine learning, computer vision, and image processing. His Ph.D. research focus is bias and self-supervised learning, especially in facial expression recognition. 
\end{IEEEbiography}

\vspace{-2.2cm}
\begin{IEEEbiography}[{\includegraphics[width=1in,height=1.25in,clip,keepaspectratio]{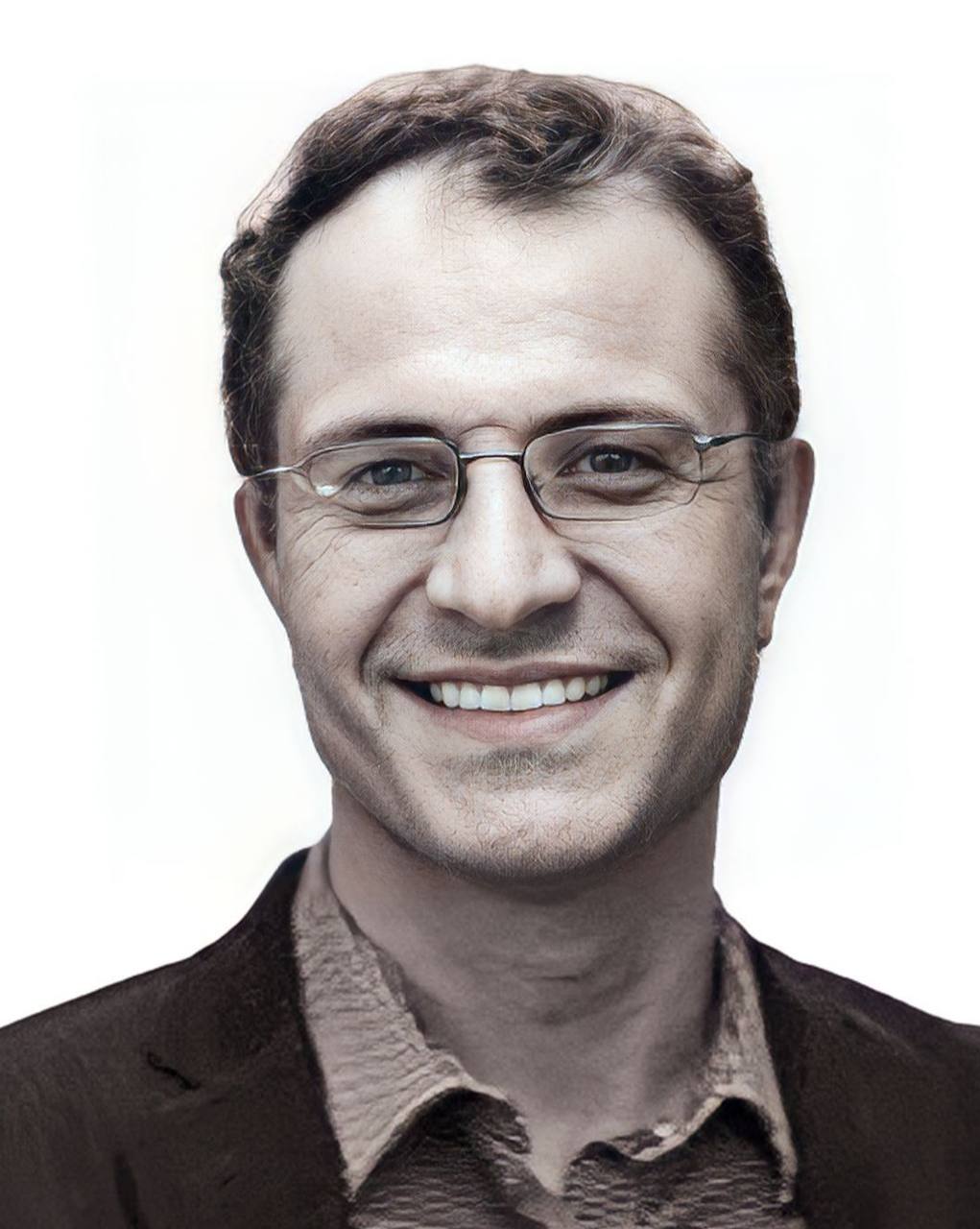}}]{Mohammad H. Mahoor} received an MS in Biomedical Engineering from Sharif University of Technology in 1998 and a Ph.D. in Electrical and Computer Engineering from the University of Miami in 2007. Currently a professor at the University of Denver, his research focuses on computer vision, deep machine learning, affective computing, and human-robot interaction, particularly with humanoid robots for children with autism and older adults with depression and dementia.
\end{IEEEbiography}

\vspace{-2.2cm}
\begin{IEEEbiography}[{\includegraphics[width=1in,height=1.25in,clip,keepaspectratio]{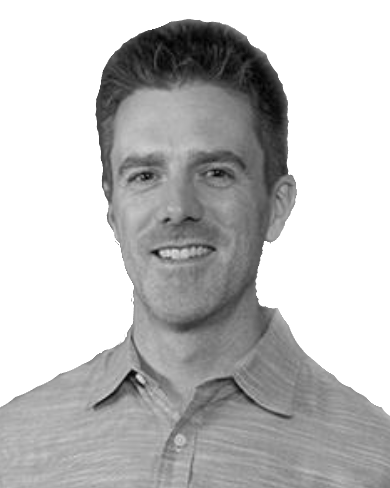}}]{Timothy Sweeny} received a Ph.D. in Psychology from Northwestern University, Evanston, Illinois, in 2010, followed by postdoctoral training at the University of California, Berkeley (2010-2013). Now an Associate Professor of Psychology at the University of Denver, he conducts research at the intersection of vision science and social psychology, focusing on visual awareness, organization, and the perception of emotion, crowds, and gaze.
\end{IEEEbiography}

\section*{Supplementary Materials}
\renewcommand{\thesection}{\Roman{section}}
\renewcommand{\thetable}{\Roman{table}}
\renewcommand{\thefigure}{\Roman{figure}}
\setcounter{section}{0}
\setcounter{table}{0}
\setcounter{figure}{0}
\begin{table}[b!] 
\caption{Review of the recent research in affective computing on some of the existing FER datasets.}
\label{tbl:fer_accuracy_supp}
\centering
\small
\resizebox{0.5\textwidth}{!}
{{
\begin{tabular}{lccc}
\hline
work                                            & Year          & Dataset            & Accuracy(\%) \\ 
\hline
Tao et al.\cite{tao2024hierarchical}            & 2024          & RAF-DB              & 91.92             \\
Li et al.\cite{li2023fg}                        & 2023          & RAF-DB              & 90.81             \\
Gong et al.\cite{gong2024enhanced}              & 2024          & Oulu-CASIA          & 89.38             \\
Sun et al.\cite{sun2023discriminatively}        & 2023          & Oulu-CASIA          & 93.34             \\
Zhao et al.\cite{zhao2022spatial}               & 2022          & Oulu-CASIA          & 89.17             \\
Gong et al.\cite{gong2024enhanced}              & 2024          & AFEW                & 53.79             \\
Savchenko et al.\cite{savchenko2022classifying} & 2022          & AFEW                & 65.50             \\
Gong et al.\cite{gong2024enhanced}              & 2024          & DFEW                & 68.78             \\
Gong et al.\cite{gong2024enhanced}              & 2024          & CK+                 & 99.04             \\
Sun et al.\cite{sun2023discriminatively}        & 2023          & CK+                 & 98.10             \\
Lee et al.\cite{lee2024hard}                    & 2024          & FER+                & 67.15             \\
Chen et al.\cite{chen2023multi}                 & 2023          & FER+                & 89.59             \\
Zhang et al.\cite{zhang2023enhanced}            & 2023          & SFEW                & 63.30             \\
Liu et al.\cite{liu2023uncertain}               & 2023          & SFEW                & 58.94             \\
Sun et al.\cite{sun2023discriminatively}	    & 2023	        & KDEF	              & 98.30	          \\
Sun et al.\cite{sun2023discriminatively}	    & 2023	        & JAFFE	              & 98.37	          \\
Cai et al.\cite{cai2022probabilistic}           & 2022          & FER-2013            & 73.28             \\
Arnaud et al.\cite{arnaud2022thin}              & 2022          & ExpW                & 76.08             \\
Cai et al.\cite{cai2022probabilistic}           & 2022          & ExpW                & 72.93             \\
Liu et al.\cite{liu2022clip}                    & 2022          & MMI                 & 91.00             \\
Zhao et al.\cite{zhao2022spatial}               & 2022          & eNTERFACE05         & 54.62             \\
Xue et al.\cite{xue2022coarse}                  & 2022          & Aff-Wild2           & 32.17             \\
Kuruvayil et al.\cite{kuruvayil2022emotion}     & 2022          & MultiPie            & 90.00             \\
Tan et al.\cite{tan2022emotion}                 & 2022          & DISFA               & 95.91             \\
Borgalli and Surve\cite{borgalli2022deep}       & 2022          & AM-FED+             & 54.13             \\
Cao et al.\cite{cao2020e2}                      & 2020          & EmotioNet           & 55.91             \\
Kartheek et al.\cite{kartheek2022windmill}      & 2022          & FERG                & 99.74             \\
Rao et al.\cite{rao2020recognition}             & 2020          & DAiSEE              & 54.42             \\
\hline 
\end{tabular}
}}
\end{table}

\begin{table}[b!] 
\caption{Review of the recent research in affective computing on AffectNet~\cite{mollahosseini2017affectnet}. Number of expressions is shown by \# Exp.}
\label{tbl:fer_accuracy_on_affectnet_supp}
\centering
\small
\resizebox{0.5\textwidth}{!}
{{
\begin{tabular}{lccc}
\hline
Work                                                    & Year                  & \# Exp  & Accuracy (\%)\\
\hline
Tao et al. \cite{tao2024hierarchical}                   & 2024                  & 7, 8       & 66.97, 63.28      \\
Chen et al. \cite{chen2023multi}                        & 2023                  & 7, 8       & 66.31, 62.48      \\
Li et al.\cite{li2023fg}                                & 2023                  & 7, 8       & 64.91, 60.69      \\
Lang et al. \cite{lang2022multi}                        & 2022                  & 7, 8       & 66.56, 63.30      \\
Wang et al. \cite{wang2022bias}                         & 2022                  & 7, 8       & 64.45, 60.24      \\
Zheng et al. \cite{zheng2022poster}                     & 2022                  & 7, 8       & 67.31, 63.34      \\
Ma et al. \cite{ma2022relation}                         & 2022                  & 7, 8       & 65.65, 61.14      \\
Lee et al.\cite{lee2024hard}                            & 2024                  & 7       & 65.29       \\
Zhang et al.\cite{zhang2023enhanced}                    & 2023                  & 8       & 61.25       \\
Liu et al.\cite{liu2023uncertain}                       & 2023                  & 8       & 62.28       \\
Liu et al. \cite{liu2023joint}                          & 2023                  & 8       & 56.80       \\
Gao et al. \cite{gao2023ssa}                            & 2023                  & 7       & 65.78       \\
Fard et al. \cite{fard2022ad}                           & 2022                  & 7       & 63.36       \\
Arnaud et al.\cite{arnaud2022thin}                      & 2022                  & 7       & 63.79       \\
Zhang et al. \cite{zhang2022improving}                  & 2022                  & 7       & 62.10       \\
Gera et al. \cite{gera2022cern}                         & 2022                  & 7       & 62.06       \\
Kuruvayil et al. \cite{kuruvayil2022emotion}            & 2022                  & 5       & 68.00       \\
Zeng et al. \cite{zeng2022face2exp}                     & 2022                  & 7       & 64.23       \\
Liu et al. \cite{liu2022uncertain}                      & 2022                  & 7       & 61.57       \\
Su et al. \cite{su2022using}                            & 2022                  & 8       & 58.68       \\
Heidari et al. \cite{heidari2022learning}               & 2022                  & 8       & 60.02       \\
\hline
\end{tabular}
}}
\end{table}
\begin{table}[b!] 
\caption{Review of the existing FER datasets, and their attributes. The symbols are as follow: I$\rightarrow$Image, VS$\rightarrow$Video Sequence,
Exp$\rightarrow$Number of Expressions, AU$\rightarrow$Action Unit, C$\rightarrow$Controlled, P$\rightarrow$Posed, S$\rightarrow$Spontaneous, W$\rightarrow$Wild, V$\rightarrow$Valence, A$\rightarrow$Arousal, MD$\rightarrow$Metadata, MA$\rightarrow$Manually Annotated, AA$\rightarrow$Automatically (Machine) Annotated, CP$\rightarrow$Compound-Label, NIR$\rightarrow$Near Infrared, D$\rightarrow$Dimension, K$\rightarrow$Kilo, M$\rightarrow$Million.}
\label{tbl:fer_datasets_supp}
\centering
\small
\resizebox{0.5\textwidth}{!}
{
\begin{tabular}{ll}
\hline
Name & Attributes \\
\hline
AffectNet\cite{mollahosseini2017affectnet} & I: $\sim$1M, Exp: 8, W, V, A, MD, MA: $\sim$440K \\ 
RAF-DB\cite{li2017reliable} & I: $\sim$30K, Exp: 19, W, MA, MD, CP  \\
CK+\cite{lucey2010extended} & VS: 593, Exp: 7, C, P, AU: 30, MA: 327  \\
Aff-Wild\cite{zafeiriou2016facial} & VS: 500, I: 10K, AU: 16, S, W, V, A  \\
Aff-Wild2\cite{kollias2018aff} & VS: 260 (+ Aff-Wild), W, V, A \\
FER-Wild\cite{mollahosseini2016facial} & I: $\sim$120K, Exp: 7, W, MD, MA: 24K  \\
MultiPie\cite{GROSS2010807} & I: $\sim$750K, Exp: 6, C, P  \\
MMI\cite{pantic2005web} & VS + I: $\sim$1.5K, Exp: 6, AU: 31, C, P, MD \\
DISFA\cite{mavadati2013disfa} & VS: 27, AU: 12, C, S, MD \\
RECOLA\cite{ringeval2013introducing} & VS: 46, Exp: 5, S, V, A, MD \\
AM-FED\cite{mcduff2013affectiva} & VS: 242, AU: 16, S, MD \\
DEAP\cite{koelstra2011deap} & VS: 32, C, S, V, A, MD  \\
AFEW\cite{dhall2012collecting} & VS:1426, Exp: 7, W, MD \\
SFEW\cite{dhall2011static} & I: 700, Exp: 7, W, MD  \\
FER-2013\cite{goodfellow2013challenges} & I: $\sim$36K, Exp: 7, W \\
EmotioNet\cite{fabian2016emotionet} & I: 1M, Exp: 23, AU: 15, W, AA, CP \\
FERG\cite{aneja2017modeling} & I: $\sim$56K, Exp: 7, Synthesized Cartoon Images  \\
Oulu-CASIA\cite{zhao2011facial} & I: $\sim$3K, Exp: 6, C, P, NIR \\
AR Face\cite{martinez1998ar} & I: $\sim$4K, Exp: 4, C, P \\
JAFFE\cite{lyons1998coding} & I: 219, Exp: 7, C, P, Japanese Females \\
GFT\cite{girard2017sayette} & VS: 96, AU: 20, C, S \\
B4PD\cite{zhang2013high} & VS: 41, Exp: 6, AU: 32, C, S, MD, 2D/3D \\
B4PD+\cite{zhang2016multimodal} & VS:140, AU: 32, C, S, MD, 2D/3D, NIR\\
4DFAB\cite{cheng20184dfab} & I: 1.8M Mesh, Exp: 6, P, S, MD, 3D/4D  \\
Belfast\cite{sneddon2011belfast} & VS: 1400, Exp: 3-5-7, C, S, V, MD \\
DAiSEE\cite{gupta2016daisee} & VS: $\sim$9K, Exp: 4, W, CP \\
FER+\cite{BarsoumICMI2016} & I: $\sim$36K, Exp: 8, W, CP \\
ExpW\cite{zhang2018facial} & I: $\sim$90K, Exp: 7, W \\
FEAFA+\cite{gan2022feafa+} & VS: 150, AU: 24, C, P, S, W  \\
KDEF\cite{lundqvist1998karolinska} & I: 490, Exp: 7, C, P, A \\
C-EXPR\cite{kollias2023multi} & VS: 400, Exp: 13, AU: 17, W, V, A, MA, MD \\
MAFW\cite{liu2022mafw} & VS: 10K, Exp: 43, W, MA, MD \\
CFEE\cite{du2014compound} & I: $\sim$5K, Exp: 22, AU: 17, C, P, AA, CP \\
iCV-MEFED\cite{guo2018dominant}  & I: $\sim$30K, Exp: 50, C, P, MA, CP \\

\hline
\end{tabular}}
\end{table}

\section{Detail on FER Methods and Datasets}
\label{SEC:DETAIL_ON_FER_METHODS_AND_DATSETS}
This section provides information regarding some of the recent proposed methods in FER and reviews the existing FER datasets. In Table~\ref{tbl:fer_accuracy_supp}, we reviewed the SOTA methods in FER and reported their accuracy and the dataset they used. As this table shows, the overall accuracies reported on the controlled datasets were higher than those reported on the wild datasets. This indicates that how real-life conditions make FER more challenging. Moreover, Table~\ref{tbl:fer_accuracy_on_affectnet_supp} demonstrates the recent research on the AffectNet~\cite{mollahosseini2017affectnet} dataset and the detail of the experiments, including year, number of expressions, and their accuracy. Likewise, Table~\ref{tbl:fer_datasets_supp} investigates the existing FER datasets and their attributes. In this table, we reported the belongings of the data in a dataset to images/videos, controlled/wild, posed/spontaneous, manual/automatic annotated, and single-label/compound-label attributes. More information, such as valence-arousal and metadata, is provided in this table.  
\section{Multi-Annotated-Set (MAS)}
\label{SEC:MULTI_ANNOTATED_SET}
\begin{figure}[t]
\centering
\caption{Distribution of images in the multi-annotated-set (MAS). Images are annotated by more than one annotator in the MAS.}
\includegraphics[width=0.35\textwidth]{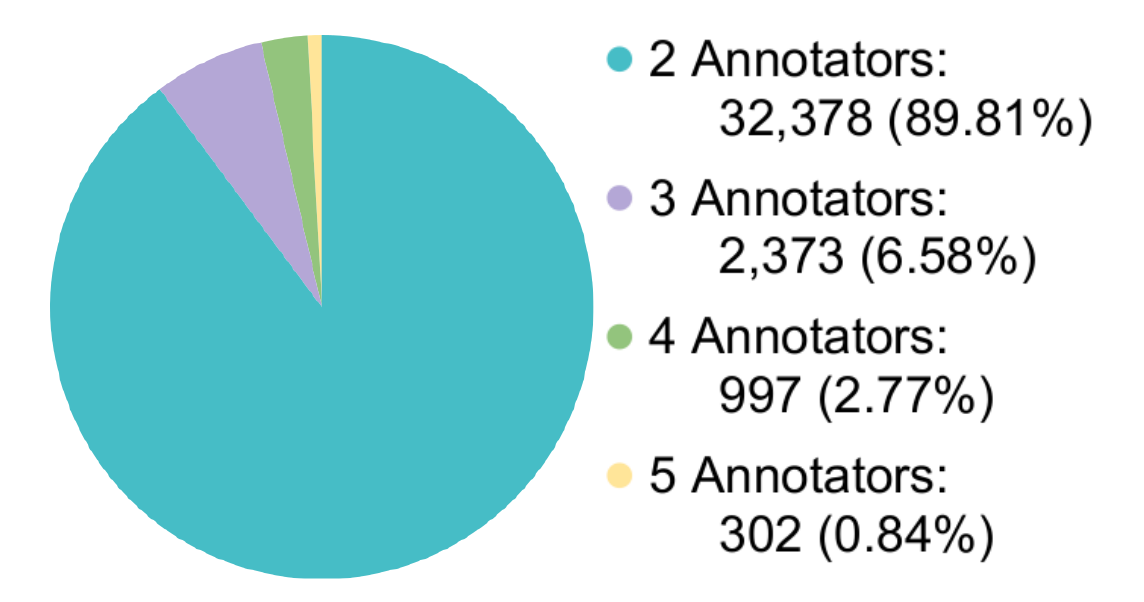}
\label{fig:mas_annotator_image}
\end{figure}

AffectNet contains a non-released subset, referred to as \textit{multi-annotated-set} (MAS), which includes more than 36K images. The images are annotated by at least two, and at most five, well trained human annotators. Since every image within the MAS is annotated with at least two well-trained human annotators, this subset is less noisy compared to the AffectNet public subsets. To be more detailed, 25 annotators were hired to annotate these images. As Fig.~\ref{fig:mas_annotator_image} shows, from all the 36,050 images, 32,378 samples are annotated by two annotators, 2,373 images have three annotators, 997 images have four annotators, and finally, five annotators labeled 302 images. Overall, there exist 76,978 annotations regarding the MAS. 

Eight emotion classes in the MAS are Neutral, Happy, Sad, Surprise, Fear, Disgust, Anger, and Contempt, accompanied by three non-expression labels, including None, Uncertain, and Non-Face. None label expresses that the facial expression of the corresponding image was none of the eight emotions. Uncertain means the annotator was uncertain of the facial expression. Likewise, Non-Face indicates that the corresponding image was not a human facial image. Another remarkable point about the MAS is labeling expression to an image based on the majority voting between annotators. When there was a tie, \textit{e.g.}, two annotators labeled an image, one as Happy and the other as Surprise, the final label was selected as the keyword used for querying that image on the web. For more details on image collection and annotation process over AffectNet refer to~\cite{mollahosseini2017affectnet}.

Since the MAS is annotated by more than one annotator, it is more reliable than the publicly available training and validation sets of AffectNet. Hence, to increase the performance of our proposed models, we used the MAS for training and test purposes. We split the MAS subset into two subsets in order to train and test our models. We first created the test set, which we refer to as test-MAS, by randomly selecting 100 images from the MAS for each emotion. The only restriction we imposed for choosing images within the test-MAS was that all the annotators agreed on a specific facial expression. Hence, test-MAS was the most reliable subset, containing images with the most clear (least ambiguous) facial expressions. Based on this clarity, we used test-MAS to assess the accuracy and performance of our proposed models. Then, we chose the rest as the training set and called it train-MAS. These two sets were \textit{only} used for the training and testing of our proposed ensemble of binary classifiers (EBC) and AU-based classifier.

\begin{table}[b!] 
\caption{Accuracy and average accuracy of Hard-FER on secondary baseline model (EfficientNet-B3\cite{tan2019efficientnet}) over AffectNet+.}
\label{tbl:efficientnet_hard_label_accuracy_supp}
\centering
\small
\resizebox{0.44\textwidth}{!}
{{
\begin{tabular}{lcccc}
\hline
        & All   & Normal & Challenging & Hard  \\ \hline
Acc (\%)     & 55.17 & 87.06  & 57.13       & 41.25 \\
$\overline{Acc}$ (\%) & 55.13 & 81.44  & 55.71       & 42.36 \\ \hline
\end{tabular}}}
\end{table}
\begin{table}[b!] 
\caption{Weighted failure-rate (W-FR), and weighted mean average error (W-MAE) of Soft-FER secondary baseline model (EfficientNet-B3\cite{tan2019efficientnet}) on AffectNet+.}
\label{tbl:efficientnet_soft_label_error_supp}
\centering
\small
\resizebox{0.45\textwidth}{!}
{{
\begin{tabular}{lcccc}
\hline
                    & All       & Normal    & Challenging & Hard  \\ \hline
W-FR (\%)           & 10.58     & 6.28      & 11.25       & 18.35 \\
W-MAE (\%)          & 17.03     & 15.13     & 18.63        & 20.81  \\
\hline
\end{tabular}}}
\end{table}
%

\section{Additional Experimental Results}
\label{SEC:ADDITIONAL_EXPERIMENTAL_RESULTS}
We utilized EfficientNet-B3~\cite{tan2019efficientnet} as our secondary baseline model. In this section, we provided the experimental results regarding the performance of Hard-FER and Soft-FER models. Overall, the results of the secondary baseline model demonstrated better performance compared to the initial baseline model (ResNet-50~\cite{he2016deep}).
\begin{figure*}[t!]
  \centering
  \includegraphics[width=1.0\textwidth]{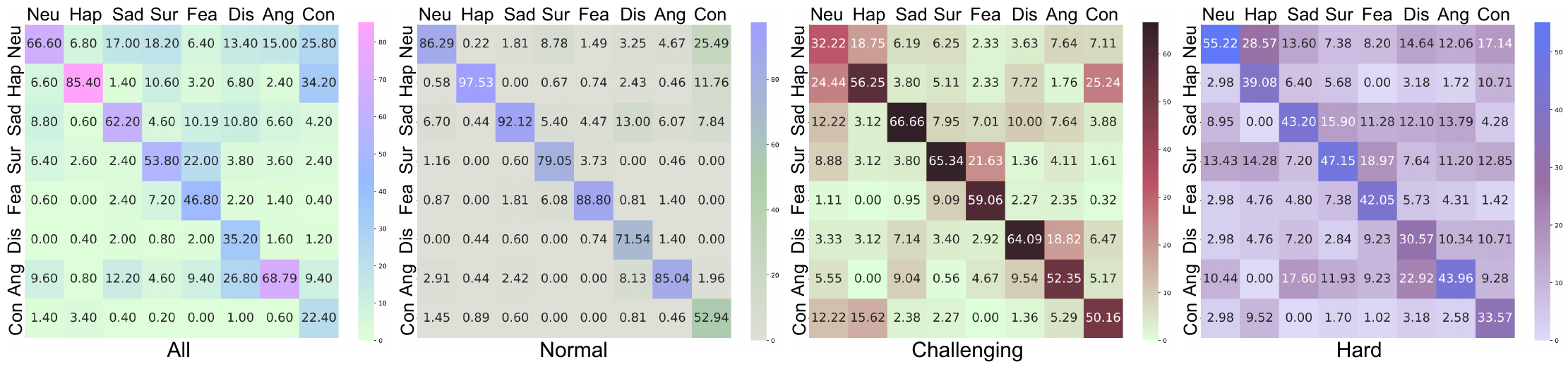}
  \caption{Confusion matrix of the secondary baseline model (EfficientNet-B3~\cite{tan2019efficientnet}) for every subset of AffectNet+.}
  \label{fig:efficientnet_confusion}
\end{figure*}
\subsection{Hard-FER Secondary Baseline}
Table~\ref{tbl:efficientnet_hard_label_accuracy_supp} shows the accuracy and average accuracy of Hard-FER secondary baseline model. This model achieved the highest accuracy (87.06\%) on the Easy subset, by far higher than the accuracy on the Challenging and Difficult subsets, 57.13\% and 41.36\%, respectively. 
Table~\ref{tbl:efficientnet_hard_label_prec_rec_f1_supp} illustrates precision, recall, and F-1 score for each subset of AffectNet+, on the secondary baseline model. EfficientNet-B3~\cite{tan2019efficientnet} reached to the highest F-1 score on the Happy expression (68.27\%), and the lowest on Contempt (34.16\%). We witnessed a similar pattern for the Easy subset, where F-1 score for Happy and Contempt was 97.16\% and 59.09\%, respectively. However, for the Challenging and Difficult subsets, the lowest F-1 score was obtained for the Neutral and Happy classes, similar to the scores reported for the main baseline model (ResNet-50~\cite{he2016deep}).
Fig.~\ref{fig:efficientnet_confusion} shows the confusion matrices of the secondary baseline model for every subset of AffectNet+. Similar to the baseline model, the secondary baseline model achieved the least confusion on the Easy subset, while the highest level of confusion occurred on the Difficult subset. 
\begin{table}[b!] 
\caption{Per-class W-FR, and W-MAE of Soft-FER secondary baseline model (EfficientNet-B3 \cite{tan2019efficientnet}) on AffectNet+.}
\label{tbl:efficientnet_soft_label_error_on_expression_supp}
\centering
\small
\resizebox{0.49\textwidth}{!}
{{
\begin{tabular}{llcccc}
\hline
                          &     & All   & Normal & Challenging & Hard  \\ \hline
\multirow{3}{*}{Neutral}  & FR  & 11.00 & 4.66   & 20.00       & 20.90 \\
                          & MAE & 17.24 & 15.09  & 22.00       & 22.42 \\
                          \hdashline
\multirow{3}{*}{Happy}    & FR  & 3.40  & 1.34   & 6.25        & 19.05 \\
                          & MAE & 11.98 & 10.79  & 18.27       & 23.02 \\
                          \hdashline
\multirow{3}{*}{Sad}      & FR  & 11.80 & 6.06   & 9.05        & 16.80 \\
                          & MAE & 17.88 & 17.62  & 17.47       & 19.64 \\
                          \hdashline
\multirow{3}{*}{Surprise} & FR  & 12.20 & 11.49  & 14.20       & 18.75 \\
                          & MAE & 17.77 & 17.20  & 19.17       & 20.41 \\
                          \hdashline
\multirow{3}{*}{Fear}     & FR  & 14.80 & 11.19  & 15.20       & 19.49 \\
                          & MAE & 18.74 & 17.81  & 19.18       & 21.75 \\
                          \hdashline
\multirow{3}{*}{Disgust}  & FR  & 12.40 & 13.01  & 10.91       & 19.11 \\
                          & MAE & 18.27 & 17.56  & 19.08       & 21.58 \\
                          \hdashline
\multirow{3}{*}{Anger}    & FR  & 11.60 & 9.35   & 11.18       & 13.79 \\
                          & MAE & 17.90 & 16.92  & 19.23       & 19.73 \\
                          \hdashline
\multirow{3}{*}{Contempt} & FR  & 7.40  & 3.92   & 7.12        & 19.29 \\
                          & MAE & 16.45 & 18.90  & 17.22       & 20.00 \\
                          \hline
\end{tabular}
}}
\end{table}

\subsection{Soft-FER Secondary Baseline}
Table~\ref{tbl:efficientnet_soft_label_error_supp} shows W-MAE and W-FR for each subset of AffectNet+. Similar to Hard-FER, the minimum W-MAE, and W-FR belonged to the Easy subset (17.03\% and 10.58\%, respectively), while the maximum was obtained for the Difficult subset (20.81\% and 18.835\%, respectively). This information reveals the degree of complexity of each subset. 

In Table~\ref{tbl:efficientnet_soft_label_error_on_expression_supp}, we analyzed the performance of the secondary baseline model in Soft-FER. For the Easy subset, we observed the lowest W-FR and W-MAE values. The highest values of W-FR and W-MAE were reported over the Difficult subset. Considering all the samples in the validation set (marked as \textit{All} in Table~\ref{tbl:efficientnet_soft_label_error_on_expression_supp}), the baseline model best recognized the Happy expression, and worst recognized Fear and Disgust emotions. For the Easy subset, the baseline model achieved the lowest W-FR and W-MAE on Happy, Neutral, and Contempt. Contrary to Hard-FER (where the baseline model showed a high confusion rate between the Neutral and Contempt classes), Soft-FER showed better performance.
\subsection{Complementary Experiments}
We reported accuracy and average accuracy over the ensemble of binary classifiers (EBC). Table~\ref{tbl:bin_eval_test_MAS_supp} reports more detail on it, including precision, recall, F-1 score, accuracy, and average accuracy. Similarly, we reported the accuracy and average accuracy of the AU-based classifier. Here, Table~\ref{tbl:action_unit_eval_test_MAS_supp} provides complementary information regarding this classifier
. In addition, to highlight the role of the action units in our models, Table \ref{tbl:action_unit_effect_on_resnet_training_supp} makes a comparison between the accuracy of the model with and without considering AUs.
\section{Metadata Analysis}
\label{SEC:METADATA_ANALYSIS}
\begin{figure*}[p]
  \centering
\caption{Covariance matrix of the facial attributes, including expression, gender, race, and age of the training sets of AffectNet+ (in \%).}
\includegraphics[width=0.95\textwidth]{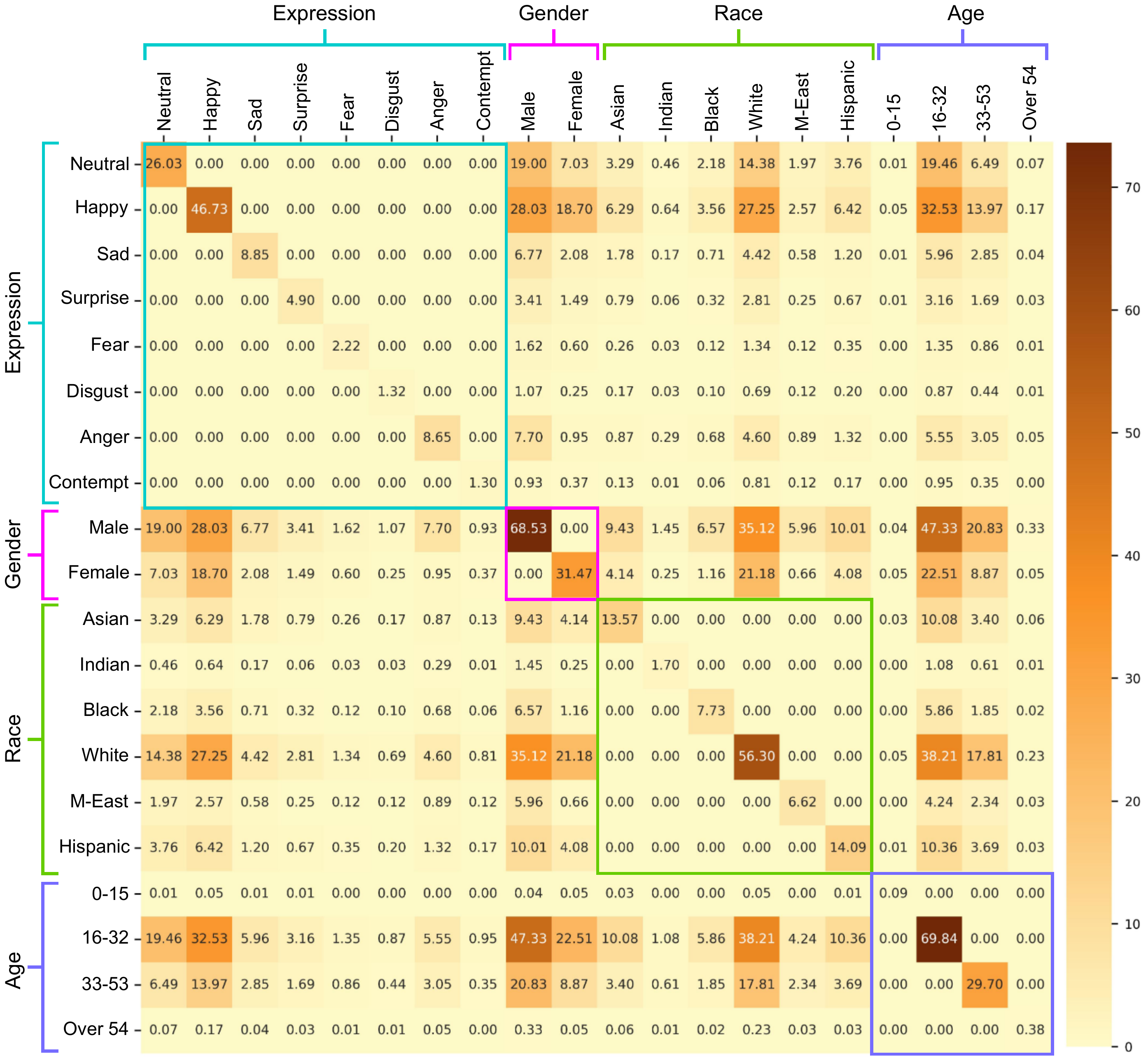}
\label{fig:features_correlation_300k}
\end{figure*}

\begin{figure*}[p]
  \centering
\caption{Covariance matrix of the facial attributes, including expression, gender, race, and age of the validation sets of AffectNet+ (in \%).}
\includegraphics[width=0.95\textwidth]{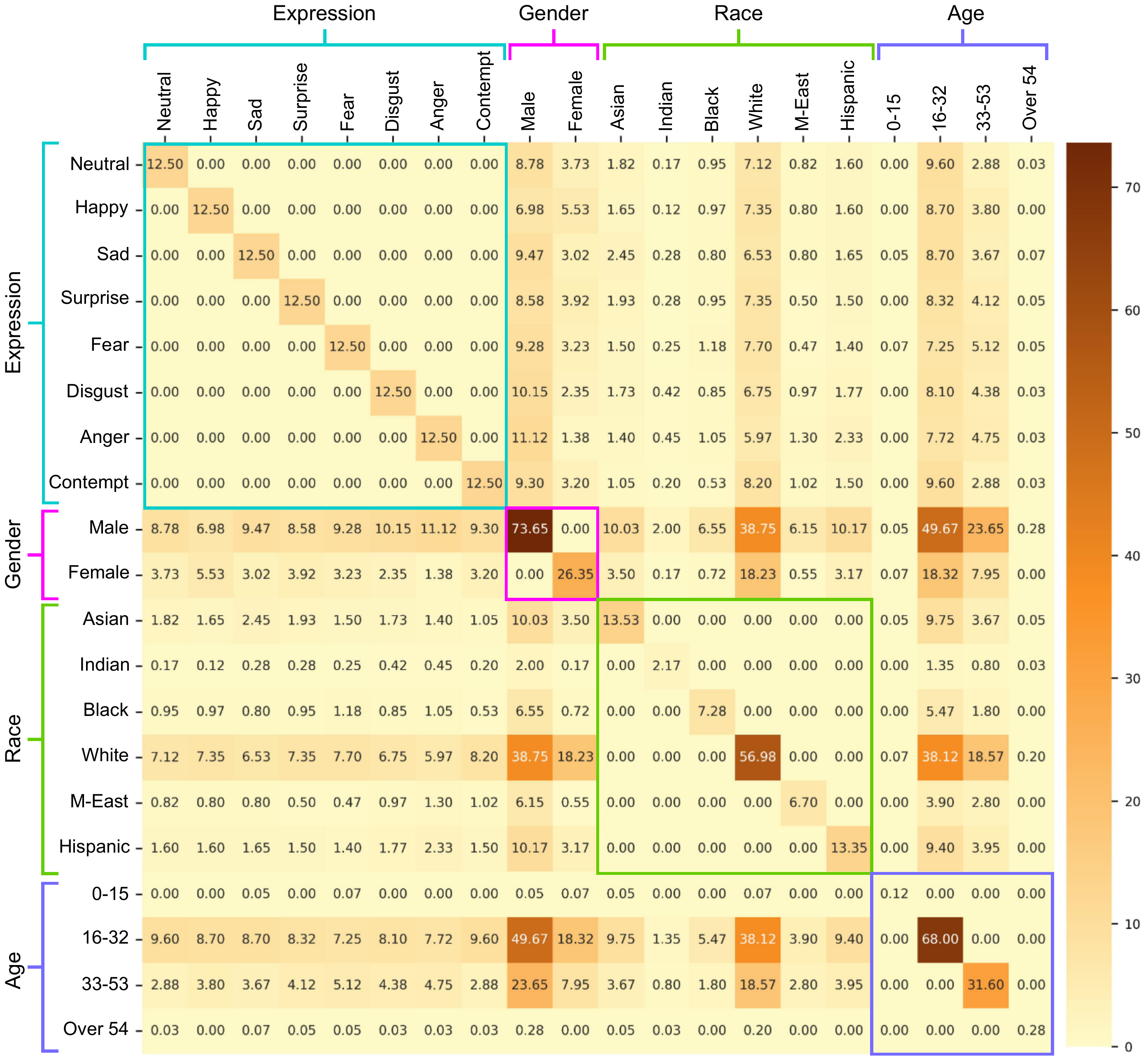}
\label{fig:features_correlation_4k}
\end{figure*}

We reported covariance matrix of the metadata and facial attributes, for training and validation sets of AffectNet+, in figures~\ref{fig:features_correlation_300k} and~\ref{fig:features_correlation_4k}, respectively. The figures show the relative proportions of different metadata and facial attributes as well as their correlation regarding the images in the training and validation sets.

As Fig.~\ref{fig:features_correlation_300k} presents, in the training set of AffectNet+, Happy included the highest portion (about 47\%) of the expression data, while Disgust and Contempt were the lowest (1.32\% and 1.30\%, respectively). Considering the gender attribute, 68.53\% of the images were categorized as Male, which was more than twice the Female images (31.47\%). Regarding the race attribute, we observed that the White race group has by far the largest portion of the data distribution, about 56.30\%. For the age attribute, the majority of 69.84\% of images were in the age range of 16-31, and 29.7\% in the 33-53 age range, while we hardly could find images belonged to the other age groups, below 16 and above 53.

Fig.~\ref{fig:features_correlation_4k} shows the covariance matrix of facial attributes for the AffectNet+ validation set. Apart from the facial expression which exhibits a balanced distribution, the remaining facial attributes in the validation set follow the same pattern in the training set.
\section{Subjective Test Detail}
\label{SEC:SUBJECTIVE_TEST_DETAIL}
\begin{figure*}[t]
\centering
\caption{Subjective test, experiments 1 and 2. In experiment 1, given an image, participants are asked to select which label type (\textit{soft-label} vs hard-label) best describe the facial expression of the image. In experiment 2, subjects should select the correct \textit{soft-label} between the correct and a randomly selected \textit{soft-label}.}
\includegraphics[width=0.95\textwidth]{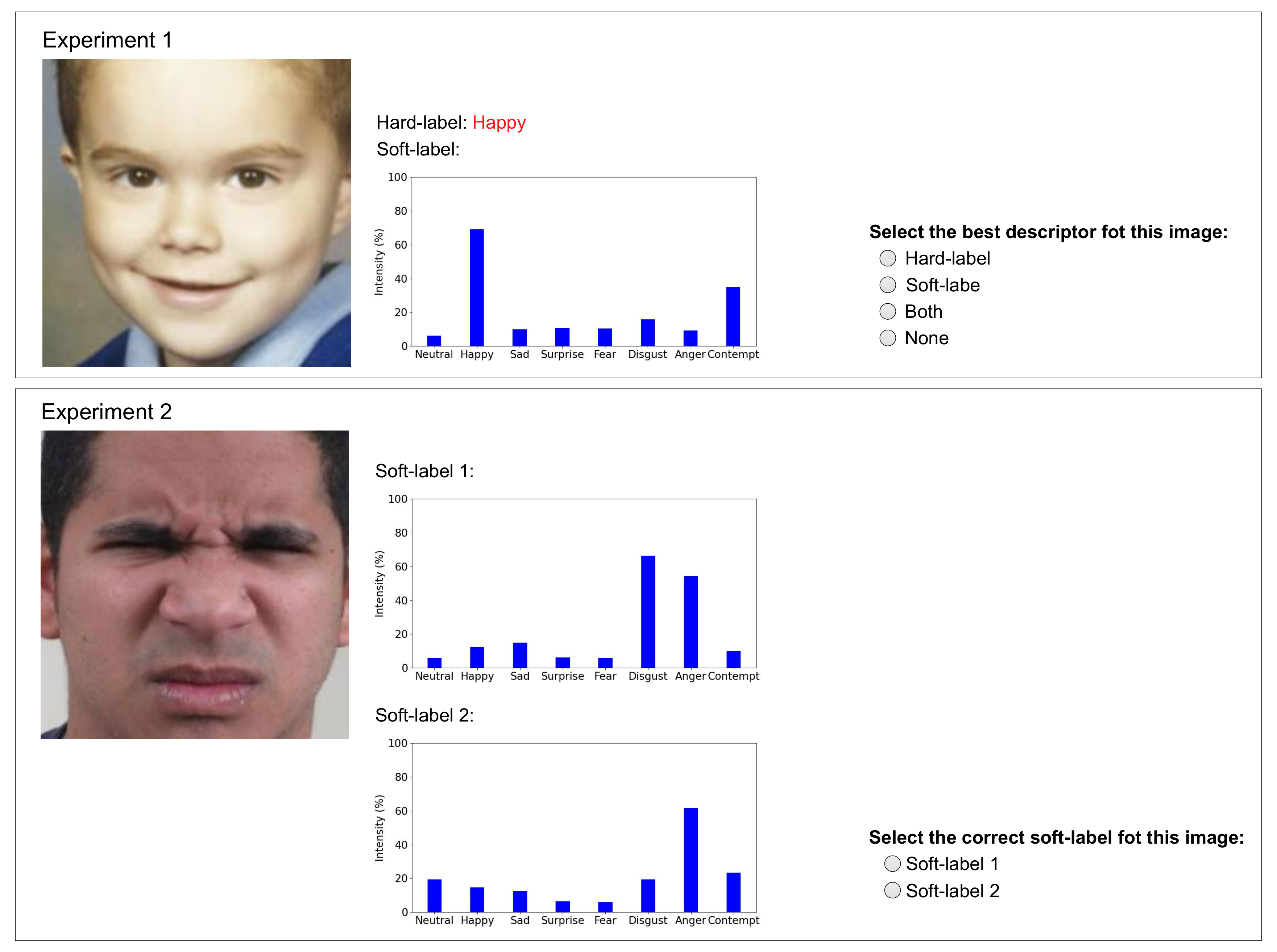}
\label{fig:ubjective_test_experiment}
\end{figure*}

In the subjective test, we asked participants 2 main questions. In the first experiment, they were responsible to choose the best image descriptor between \textit{soft-label} or hard-label, while they had also the option to choose both or none. Both means when \textit{soft-label} has only one intense column which is correctly agreed with the hard-label. None means both of the \textit{soft-label} and hard-label are incorrect. In the second experiment participants should find the correct \textit{soft-label} between two shown \textit{soft-labels}, where one of them was corresponded to the image and the other was randomly selected. It is notable that we shuffled all the experiments to avoid bias toward a question or an image. Figure~\ref{fig:ubjective_test_experiment} shows our two experiments.

\begin{table*}[p] 
\caption{Per-class precision, recall, and F-1 score of Hard-FER secondary baseline model, EfficientNet-B3\cite{tan2019efficientnet}, for each expression on the AffectNet+ dataset (in \%).}
\label{tbl:efficientnet_hard_label_prec_rec_f1_supp}
\centering
\small
\resizebox{0.55\textwidth}{!}
{{
\begin{tabular}{lcccccccc}
\hline
     & Neutral & Happy & Sad   & Surprise & Anger & Disgust & Fear  & Contempt \\ \hline
     & \multicolumn{8}{c}{\textbf{All}}                                                 \\ \hdashline
Prec & 39.35   & 56.89 & 57.83 & 55.64    & 76.41 & 81.30   & 48.64 & 75.69    \\
Rec  & 66.73   & 85.34 & 62.50 & 53.73    & 46.84 & 35.15   & 68.68 & 22.06    \\
F-1  & 49.51   & 68.27 & 60.07 & 54.67    & 58.08 & 49.08   & 56.95 & 34.16    \\ \hline
     & \multicolumn{8}{c}{\textbf{Normal}}                                              \\ \hdashline
Prec & 86.64   & 96.83 & 67.12 & 91.40    & 86.02 & 92.47   & 86.95 & 68.42    \\
Rec  & 86.13   & 97.49 & 91.87 & 79.05    & 88.63 & 71.07   & 85.30 & 52.00    \\
F-1  & 86.39   & 97.16 & 77.57 & 84.78    & 87.31 & 80.37   & 86.12 & 59.09    \\ \hline
     & \multicolumn{8}{c}{\textbf{Challenging}}                                         \\ \hdashline
Prec & 27.18   & 11.39 & 62.22 & 62.08    & 77.69 & 63.88   & 55.41 & 80.42    \\
Rec  & 31.46   & 56.25 & 66.98 & 65.31    & 59.76 & 63.88   & 52.40 & 49.67    \\
F-1  & 29.16   & 18.94 & 64.51 & 63.66    & 67.55 & 63.88   & 53.86 & 61.41    \\ \hline
     & \multicolumn{8}{c}{\textbf{Hard}}                                                \\ \hdashline
Prec & 24.64   & 16.32 & 35.86 & 44.63    & 67.82 & 43.63   & 31.44 & 73.01    \\
Rec  & 55.55   & 42.10 & 42.97 & 46.74    & 41.71 & 31.57   & 44.64 & 33.57    \\
F-1  & 34.14   & 23.52 & 39.09 & 45.66    & 51.65 & 36.64   & 36.90 & 46.00    \\ \hline
\end{tabular}
}}
\end{table*}
\begin{table*}[p] 
\caption{Precision, recall, F-1 score, accuracy and average accuracy of ensemble of binary classifiers (EBC), over test-MAS (in \%).}
\label{tbl:bin_eval_test_MAS_supp}
\centering
\small
\resizebox{0.55\textwidth}{!}
{{
\begin{tabular}{lcccccccc}
\hline
& Neutral & Happy & Sad  & Surprise & Fear & Disgust & Anger & Contempt \\ \hline
\multicolumn{9}{c}{\textbf{ResNet-50\cite{he2016deep}}} \\ \hdashline
Prec    & 67.0    & 71.0  & 65.9 & 73.0     & 73.4 & 74.9    & 68.5  & 59.6     \\
Rec     & 79.5    & 87.1  & 77.0 & 85.7     & 83.9 & 80.4    & 83.9  & 65.2     \\
F-1     & 69.7    & 74.6  & 68.3 & 76.8     & 76.9 & 77.2    & 71.2  & 60.7     \\
Acc     & 81.4    & 84.1  & 80.8 & 86.9     & 87.5 & 88.9    & 81.5  & 77.4     \\
$\overline{Acc}$ & 79.5    & 87.1  & 77.0 & 85.6     & 83.9 & 80.4    & 83.9  & 65.2     \\ 
\hline
\multicolumn{9}{c}{\textbf{EfficientNet-B3\cite{tan2019efficientnet}}} \\ \hdashline
Prec    & 69.1    & 71.4  & 67.1 & 74.4     & 72.4 & 73.3    & 70.3  & 60.8     \\
Rec     & 83.0    & 87.3  & 79.9 & 86.3     & 82.7 & 85.8    & 82.7  & 69.7     \\
F-1     & 72.1    & 75.0  & 69.7 & 78.2     & 75.9 & 77.1    & 73.6  & 61.6     \\
Acc     & 82.9    & 84.5  & 81.3 & 88.0     & 86.9 & 87.1    & 84.6  & 75.4     \\
$\overline{Acc}$ & 82.9    & 87.3  & 79.9 & 86.3     & 82.6 & 85.8    & 82.6  & 69.6     \\ 
\hline
\multicolumn{9}{c}{\textbf{XceptionNet\cite{chollet2017xception}}} \\ \hdashline
Prec    & 67.6    & 70.4  & 68.8 & 76.3     & 74.6 & 79.7    & 74.6  & 65.1     \\
Rec     & 81.6    & 88.0  & 82.8 & 87.5     & 84.8 & 87.5    & 87.6  & 65.5     \\
F-1     & 70.3    & 73.6  & 71.9 & 80.2     & 78.2 & 82.9    & 78.6  & 65.3     \\
Acc     & 81.3    & 82.8  & 82.6 & 89.4     & 88.4 & 91.5    & 88.0  & 84.6     \\
$\overline{Acc}$ & 81.6    & 88.0  & 82.8 & 87.5     & 84.8 & 87.4    & 87.6  & 65.5     \\ 
\hline
\multicolumn{9}{c}{\textbf{Ensemble of Binary Classifiers}} \\ \hdashline
Prec    & 71.6    & 71.3    & 75.1    & 77.0    & 82.2    & 80.0    &    70.9 & 73.0 \\
Rec & 88.5 & 88.0 & 84.6 & 91.1 & 88.6 & 86.7 & 85.2 & 78.5 \\
F-1     & 75.2    & 74.8  & 78.5  & 81.5     & 85.0 & 82.8    & 74.5  & 75.3     \\
Acc     & 84.4    & 84.1  & 88.8  & 89.6     & 92.8 & 91.6    & 84.6  & 87.9     \\
$\overline{Acc}$ & 88.5    & 87.9  & 84.6  & 91.1     & 88.6 & 86.6    & 85.2  & 78.5     \\ 
\hline
\end{tabular}
}}
\end{table*}

\begin{table*}[p] 
\caption{Precision, recall, F-1 score, accuracy and average accuracy of AU-based classifier, over test-MAS (in \%).}
\label{tbl:action_unit_eval_test_MAS_supp}
\centering
\small
\resizebox{0.55\textwidth}{!}
{{
\begin{tabular}{lcccccccc}
\hline
 & Neutral & Happy & Sad   & Surprise & Fear & Disgust & Anger & Contempt \\ \hline
Prec    & 71.9    & 69.7  & 61.45 & 66.8     & 57.0 & 67.4    & 57.9  & 62.6     \\
Rec     & 88.34   & 87.2  & 75.6  & 85.4     & 60.2 & 81.4    & 67.5  & 77.5     \\
F-1     & 75.6    & 72.5  & 58.5  & 67.3     & 31.4 & 69.9    & 46.3  & 60.1     \\
Acc     & 84.9    & 82.0  & 67    & 75.9     & 31.8 & 80.9    & 50.5  & 70.3     \\
$\overline{Acc}$ & 88.4    & 87.1  & 75.6  & 85.4     & 60.1 & 81.4    & 67.4  & 77.4     \\ \hline
\end{tabular}
}}
\end{table*}
\begin{table*}[t!] 
\caption{Comparison between the results of the ResNet-50~\cite{he2016deep} model with and without action units. All the results are reported in percent.}
\label{tbl:action_unit_effect_on_resnet_training_supp}
\centering
\small
\resizebox{0.75\textwidth}{!}
{{
\begin{tabular}{llcccccccc}
\hline
   &     & Neutral & Happy & Sad   & Surprise & Fear  & Disgust & Anger & Contempt \\ \hline
\multirow{2}{*}{Only Binary Model} & $Acc$ & 81.37   & 84.13 & 80.75 & 86.88    & 87.50 & 88.88    & 81.50 & 77.38    \\  
  & $\overline{Acc}$ & 79.50   & 87.07 & 77.00 & 85.64    & 83.86 & 80.36   & 83.86 & 65.21    \\ \hline
\multirow{2}{*}{Binary + AU-based Model}  & $Acc$ & 84.38   & 84.13 & 88.75 & 89.63    & 92.75 & 91.63    & 84.63 & 87.88    \\ 
  & $\overline{Acc}$ & 88.50   & 87.93 & 84.57 & 91.07    & 88.57 & 86.64   & 85.21 & 78.50    \\ \hline
\end{tabular}
}}
\end{table*}

\end{document}